\documentclass{article}

\PassOptionsToPackage{numbers, compress}{natbib}
\usepackage[preprint]{neurips_2026}

\usepackage[utf8]{inputenc} % allow utf-8 input
\usepackage[T1]{fontenc}    % use 8-bit T1 fonts
\usepackage{hyperref}       % hyperlinks
\usepackage{url}            % simple URL typesetting
\usepackage{booktabs}       % professional-quality tables
\usepackage{graphicx}
\usepackage{amsfonts}       % blackboard math symbols
\usepackage{nicefrac}       % compact symbols for 1/2, etc.
\usepackage{microtype}      % microtypography
\usepackage{xcolor}         % colors
\usepackage{float}
\usepackage{dblfloatfix}
\usepackage{listings}
\definecolor{codegreen}{rgb}{0,0.6,0}
\definecolor{codegray}{rgb}{0.5,0.5,0.5}
\definecolor{codepurple}{rgb}{0.58,0,0.82}
\definecolor{backcolour}{rgb}{0.95,0.95,0.95}
\lstdefinelanguage{PDDL}{
  sensitive=true,
  morecomment=[l]{;},                 % PDDL-style comments (if you use them)
  morestring=[b]",                    % strings in quotes
  alsoletter={:-_},                   % treat :, -, _ as part of words
  morekeywords=[1]{define,problem,domain,And,On,Turnon,CloseXY},
  morekeywords=[2]{:domain,:language,:regions,:target,:ranges,:yaw_rotation,
                   :fixtures,:objects,:obj_of_interest,:moving_objects,:init,:goal},
  keywordstyle=[1]\color{codepurple}\bfseries,
  keywordstyle=[2]\color{teal}\bfseries,
}

\lstdefinestyle{mystyle}{
    backgroundcolor=\color{backcolour},   
    commentstyle=\color{codegreen},
    keywordstyle=\color{magenta}\textbf,
    numberstyle=\tiny\color{codegray},
    stringstyle=\color{codepurple},
    basicstyle=\bfseries\ttfamily\footnotesize,
    breakatwhitespace=false,         
    breaklines=true,                 
    captionpos=b,                    
    keepspaces=true,                 
    numbers=left,                    
    numbersep=5pt,                  
    showspaces=false,                
    showstringspaces=false,
    showtabs=false,                  
    tabsize=2
}
\usepackage{makecell}
\usepackage{array}
\usepackage{wrapfig}
\usepackage{multicol,multirow}
\usepackage{enumitem}
\usepackage{amsmath}
\usepackage{amssymb}
\usepackage{mathtools}
\usepackage{amsthm}
\usepackage{subcaption}
\newcommand{\method}{DySta}
\newcommand{\bs}[1]{\boldsymbol{#1}}

\usepackage{hyperref}

\title{Static-Dynamic Disentanglement for Efficient Multi-Frame Vision-Language-Action Models}

\author{%
  Weikang Qiu\thanks{Equal contribution.} \\
  Yale University\\
  \And
  Huashuo Lei\footnotemark[1] \\
  The Hong Kong University of Science and Technology \\(Guangzhou) \\
  \AND
  Tinglin Huang \\
  Yale University\\
  \And
  Rex Ying \\
  Yale University\\
}

\begin{document}

\maketitle

\begin{abstract}

Vision-Language-Action (VLA) models have recently emerged as a promising paradigm for generalist robotic control. Built upon vision–language model (VLM) architectures, VLAs predict actions conditioned on visual observations and language instructions, achieving strong performance and generalization across tasks. However, VLAs face two major challenges: a limited context window for input frames and inefficient inference due to the quadratic attention complexity and large parameter counts. To this end, we propose \method{}, a framework that disentangles visual inputs into multi-level static and dynamic tokens, which enables (1) retaining a single copy of static tokens across frames to significantly reduce context length, and (2) reusing the key–value (KV) cache of static tokens through a lightweight recache gate that updates only when necessary. This design enables efficient multi-frame integration and efficient inference.
In addition, we introduce a new benchmark that more effectively evaluates the multi-frame integration ability of VLAs. Experiments show that \method{} improves multi-frame integration by 24.5\% across metrics on our benchmark and 23.3\% in absolute success rate on real-world memory-dependent tasks, while accelerating inference by 2.0× (with +2.3\% success rate) on simulation benchmarks and 2.2× (with +10.6\% success rate) on real-world general tasks.
\end{abstract}

\section{Introduction}

\begin{wrapfigure}[22]{r}{0.65\textwidth}
\vspace{-1em}
    \centering
    \noindent\makebox[\linewidth]{%
    \includegraphics[width=\linewidth]{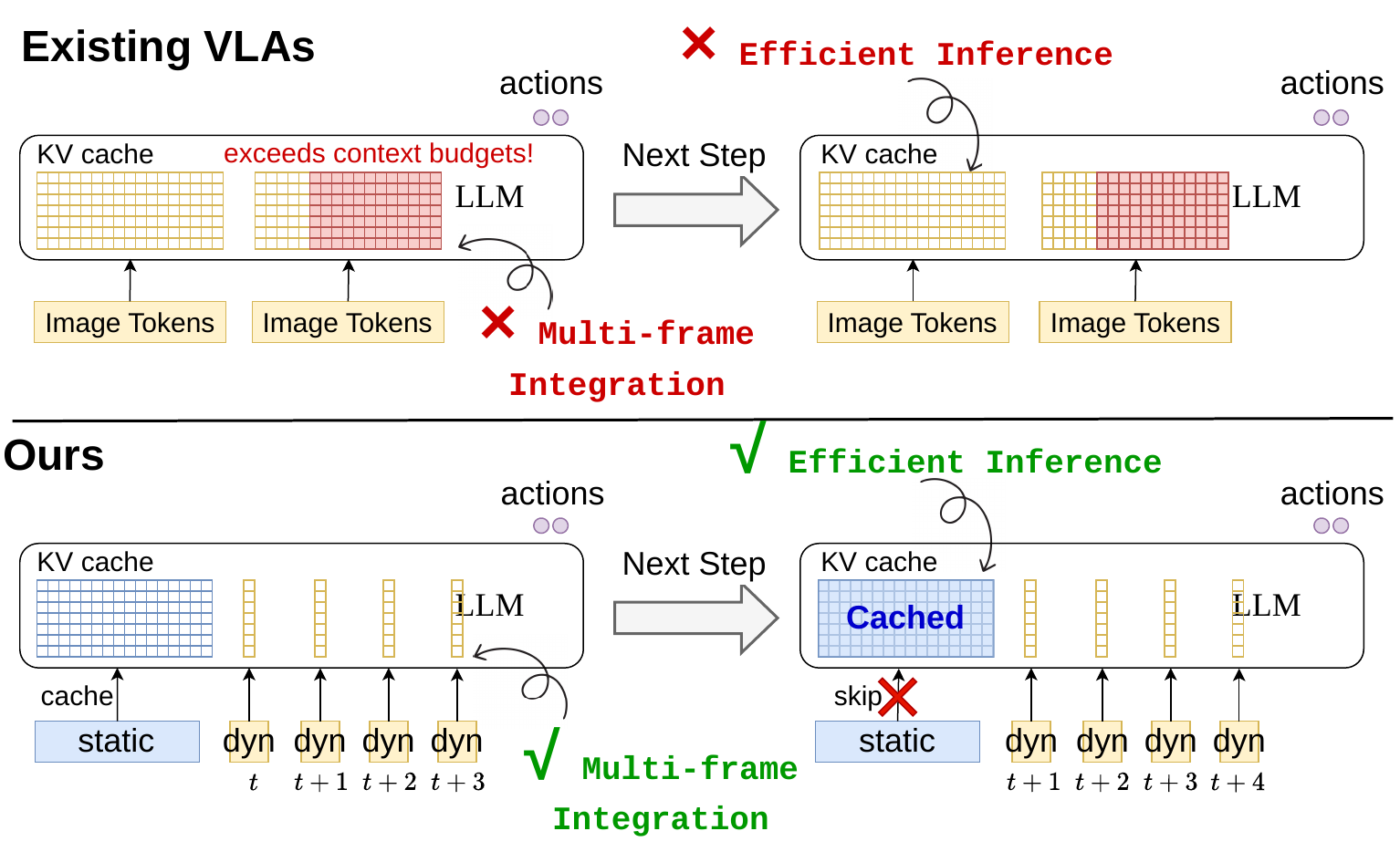}
    }
% \begin{subfigure}[b]{0.4\textwidth}
%     \centering
%     \includegraphics[height=7cm]{figs/teaser1.pdf}
%     \label{fig:0-input}
% \end{subfigure}
% \begin{subfigure}[b]{0.4\textwidth}
%     \centering
%     \includegraphics[height=7cm]{figs/teaser2.pdf}
%     \label{fig:0-input}
% \end{subfigure}
%     \caption{
%         Caption
%     }
\caption{Our model solves two main challenges in existing VLAs with the proposed static-dynamic disentanglement. 1) By keeping one copy of static tokens across all timesteps, our model is able to squash observations of multiple steps to the model's context; 2) By moving static tokens in front of all dynamic tokens, our model could reuse the KV-cache of previous timesteps during rollouts. }
    \label{fig:intro}
\end{wrapfigure}

Vision-Language-Action (VLA) models \cite{li2024cogact,kim2024openvla,qu2025spatialvla,kim2025fine,zhu2024surveymodelcompressionlarge,zhong2025surveyvisionlanguageactionmodelsaction,black2024pi0visionlanguageactionflowmodel,intelligence2025pi05visionlanguageactionmodelopenworld} have recently emerged as a powerful paradigm for generalist robotic control. Trained on large-scale heterogeneous datasets such as Open-X-Embodiment \cite{vuong2023open}, which aggregate expert demonstrations across diverse robotic tasks, state-of-the-art VLA models exhibit impressive task performance. However, they are typically built on top of large vision–language models \cite{liu2023visual,karamcheti2024prismatic,beyer2024paligemma,steiner2024paligemma}, whose large parameter counts and quadratic-complexity contextual processing \cite{vaswani2017attention} translate into two central deployment bottlenecks: (1) a constrained context budget that limits how many past observations can be ingested, and (2) high per-step inference latency that hinders real-time control. These manifest as the two challenges of \emph{multi-frame integration} and \emph{efficient inference} elaborated below and illustrated in Figure~\ref{fig:intro}.

\noindent\textbf{Multi-frame Integration}. Most current VLAs operate in a memoryless manner \cite{kim2024openvla,li2024cogact,qu2025spatialvla}, taking only the current observation as input. As a result, they struggle with tasks that require temporal dependency or memory tracking. For instance, when instructed to press a button, a VLA must remember whether the button has already been pressed; otherwise, it may repeat the same action indefinitely. A straightforward solution is to include previous observations in the model input. However, VLA's vision backbones typically produce hundreds of visual tokens per image, leading to prohibitively long contexts when multiple frames are concatenated for the transformer-based architecture with quadratic complexity in sequence length. Existing methods often rely on indirect or compressed representations \cite{jang2025contextvla,torne2026mem}, or expose multiple frames to the decoder head only \cite{shi2025memoryvlaperceptualcognitivememoryvisionlanguageaction}, which either risks significant information loss or bypasses the language model’s ability to jointly reason over multiple frames.

\noindent\textbf{Efficient Inference}. Due to their large model sizes, VLAs incur long latency for each forward pass. In real-world settings, however, robotic systems are often expected to respond promptly—for example, a household assistant should complete tasks as quickly as possible, while safety-critical scenarios such as spill containment or fire response may require near-instantaneous reactions. Moreover, recent post-training approaches for VLAs \cite{li2025simplevlarl,lu2025vla}, rely on reinforcement learning \cite{sutton1999policy,schulman2017proximal,shao2024deepseekmath} and require extensive rollouts during training, making inference speed a key bottleneck. While existing work has explored improving VLA efficiency through generic techniques for general machine learning models, such as quantization, token pruning, or layer pruning \cite{xu2025vla,yue2024deer,yang2025efficientvla,kim2024openvla}, these approaches do not leverage the intrinsic characteristics of VLA tasks. Recent works \cite{xu2025vla,liu2025ttf,tan2025think} exploit temporal redundancy by reusing computations across consecutive frames, via KV-cache reuse or action reuse. Nonetheless, these techniques rely on heuristic, non-learnable criteria and often assume that visual similarity in pixel space indicates the temporal-consistency of their latent representations—an assumption that is invalid in transformer-based vision and language backbones, as illustrated in Figure~\ref{fig:related_work1}.

To address these challenges, we propose \method{}. We focus on a broad and practically important class of robotic scenarios—such as tabletop manipulation, household chores, and warehouse picking—where the agent operates within relatively stable scenes or undergoes discrete, infrequent scene transitions. In this regime, much of the visual information in a scene remains static or changes slowly over time: the background, objects remaining still, and even moved objects whose visual appearances stay largely invariant. Building on this observation, we explicitly disentangle visual tokens into dynamic tokens and multi-level static tokens with different temporal persistence. This design yields two key benefits. First, instead of conditioning on a single image as in prior work, our model can ingest multi-step observations while maintaining a compact context: static tokens are included only once in the input sequence across timesteps, while only dynamic tokens from multiple steps are concatenated over time. This substantially reduces the effective context length and enables multi-frame integration. Second, our method improves the inference efficiency by reusing key–value (KV) caches associated with static tokens from previous steps. We also introduce a recache gate module that determines whether previously cached static tokens should be reused or recomputed, which further improves performance while minimizing the inference latency.

In addition, we observe that existing simulation benchmarks largely fail to assess a VLA's multi-frame integration capabilities. For example, temporally dependent tasks such as placing objects into a basket do not require remembering past trajectories. To address this gap, we design a new simulation benchmark inspired by principles of human episodic memory, which provides a more effective evaluation of temporal reasoning and memory usage than prior benchmarks \cite{liu2023libero,li2024evaluating}.

To evaluate multi-frame integration, we assess our model on the proposed benchmark. Our method achieves a $24.5\%$ improvement across all metrics and $23.3\%$ in absolute success rate on real-world experiments. To evaluate inference efficiency, we benchmark our approach on public simulation benchmarks~\cite{liu2023libero,li2024evaluating} and real-world experiments. On SimplerEnv~\cite{li2024evaluating}, \method{} improves the success rate by $3.9\%$ over the strongest baseline while achieving a $2.26\times$ inference speedup. In real-world experiments, \method{} improves the average success rate by $10.6\%$ over the base model and achieves a $2.21\times$ speedup. These results demonstrate that \method{} not only enhances multi-frame integration, but also substantially reduces inference latency. In summary, our contributions are as follows:

\begin{itemize}[noitemsep,topsep=0pt,leftmargin=*]
    \item We propose \method{}, which enables multi-frame integration and efficient inference by disentangling image tokens to dynamic tokens and multi-level static tokens with different temporal persistence.
    \item We introduce a trainable recache gate that adaptively determines when to refresh the cache or reuse previously cached representations, further improving the performance while minimizing the inference latency.
    \item We introduce LIBERO-Memory, a new benchmark that more effectively evaluates a VLA’s multi-frame integration ability by temporally dependent tasks.
\end{itemize}
\section{Related Works}

\noindent\textbf{Vision-Language-Action Models}
Vision-language models (VLMs) \cite{liu2023visual,wang2024cogvlm} have demonstrated strong performance in image-related and cross-modal tasks. Leveraging these capabilities, vision-language-action (VLA) models further finetune VLMs on large-scale robotic datasets \cite{vuong2023open,liu2023libero}, which usually consist of extensive human demonstrations. At each timestep, VLAs \cite{black2024pi0visionlanguageactionflowmodel,intelligence2025pi05visionlanguageactionmodelopenworld,kim2024openvla,li2024cogact,qu2025spatialvla} usually take the current environment observation (image) and a language instruction as inputs, and are trained to output actions of the next step. Although they have demonstrated high performance on certain benchmarks, due to the large parameters, they are 1) inefficient in rollout and 2) unable to incorporate historical frames in the context. In contrast, our model, by disentangling the static and dynamic components, is efficient in inference by leveraging the KV-cache of static components, and is able to incorporate a long history for temporally dependent tasks. 
\begin{wrapfigure}[19]{r}{0.5\textwidth}
    \centering
    \includegraphics[width=\linewidth]{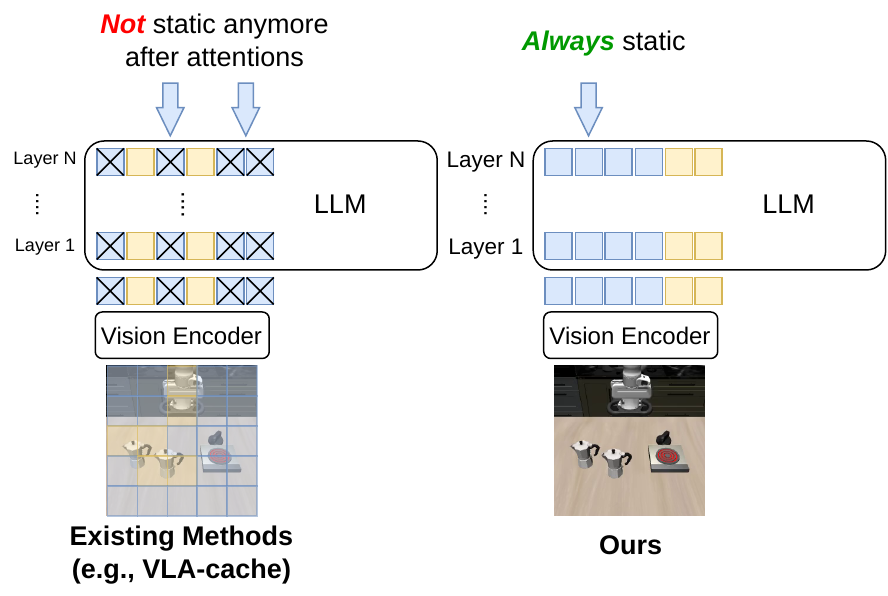}
    \caption{Some of the existing methods \cite{xu2025vla,liu2025ttf} that exploit to reuse information in previous frames ignore the problem that the static patches will be affected by attentions even if they are identical in the original pixel space.}
    \label{fig:related_work1}
    % \vspace{-1em}
\end{wrapfigure}

\noindent\textbf{VLA Acceleration}
Methods to improve VLA's efficiency usually focus on general quantization or pruning methods. For example, \citep{kim2024openvla,reddy2019sqilimitationlearningreinforcement,park2024quantizationawareimitationlearningresourceefficientrobotic} explores quantization techniques applied to VLAs. \citep{shukor2025smolvlavisionlanguageactionmodelaffordable,yang2025efficientvla,zhang2025mole,yue2024deer} explored pruning unimportant tokens and layers according to some heuristics. Although reducing the computational complexity of VLAs, these methods just migrate acceleration methods from existing general machine learning models, thus totally ignoring the unique characteristics of VLAs. Different from these methods, recent works \cite{xu2025vla,liu2025ttf,tan2025think} explore temporal redundancy in decision making: consecutive frames share temporal correlation, enabling avoiding recomputation of some components by, for example, leveraging the KV-cache of previous steps. However, these methods rely on non-learnable heuristics to recache or simply reuse the previous action, which may suffer from suboptimal performance. Notably, some of these methods \cite{xu2025vla,liu2025ttf} implicitly assume that visual similarity in pixel space implies invariance in the latent representations produced by the vision encoder and LLM—an assumption that does not generally hold in transformer-based architectures, as illustrated in Figure~\ref{fig:related_work1} and elaborated in Appendix~\ref{app:drift}.

\noindent\textbf{Multi-Frame VLAs}
Existing approaches to incorporating historical frames in VLAs typically provide the model with only indirect or compressed access to temporal information. For example, \citet{jang2025contextvla} uses a non-learnable pooling operation to condense historical frames into a fixed number of tokens. \citet{liu2025ttf,torne2026mem} incorporate historical frames into the current input via patch-wise mixing. TraceVLA \cite{zheng2024tracevla} overlays visual traces of keypoints onto the current observation to indicate object trajectories. In addition, MemoryVLA \cite{shi2025memoryvlaperceptualcognitivememoryvisionlanguageaction} exposes the LLM backbone to only a single frame at each timestep and delegates multi-frame reasoning to a lightweight decoding module, which limits the model’s capacity for joint temporal decision-making. In contrast to these methods, our approach allows the LLM backbone to directly reason over multiple frames without information loss, by explicitly disentangling static and dynamic components and avoiding redundant involvement of temporally persistent visual information.
\section{Method}
In this section, we first present the problem formulation. We then introduce the architecture of our model, followed by the training objectives that enable static–dynamic disentanglement and the recaching mechanism. Next, we provide a computational complexity analysis and derive the theoretical acceleration and context-length improvements. Finally, we introduce the benchmark for evaluating multi-frame integration.

\noindent\textbf{Problem Formulation}
Given a timestep $t$, the corresponding observation of the environment is denoted by $\bs{X}_t$. A VLA $\pi$ predicts actions of the following steps based on previous $T$ observations and a language instruction $\mathcal{I}$.
\begin{equation}
    \pi: (\bs{X}_{t-T}, \bs{X}_{t-T+1}, \cdots, \bs{X}_{t}, \mathcal{I}) \mapsto \bs{a}_t, \bs{a}_{t+1}, \cdots,
\end{equation}

\begin{wrapfigure}[20]{r}{0.5\textwidth}
\vspace{-2em}
    \centering
    \noindent\makebox[\linewidth]{%
    \includegraphics[width=1.0\linewidth]{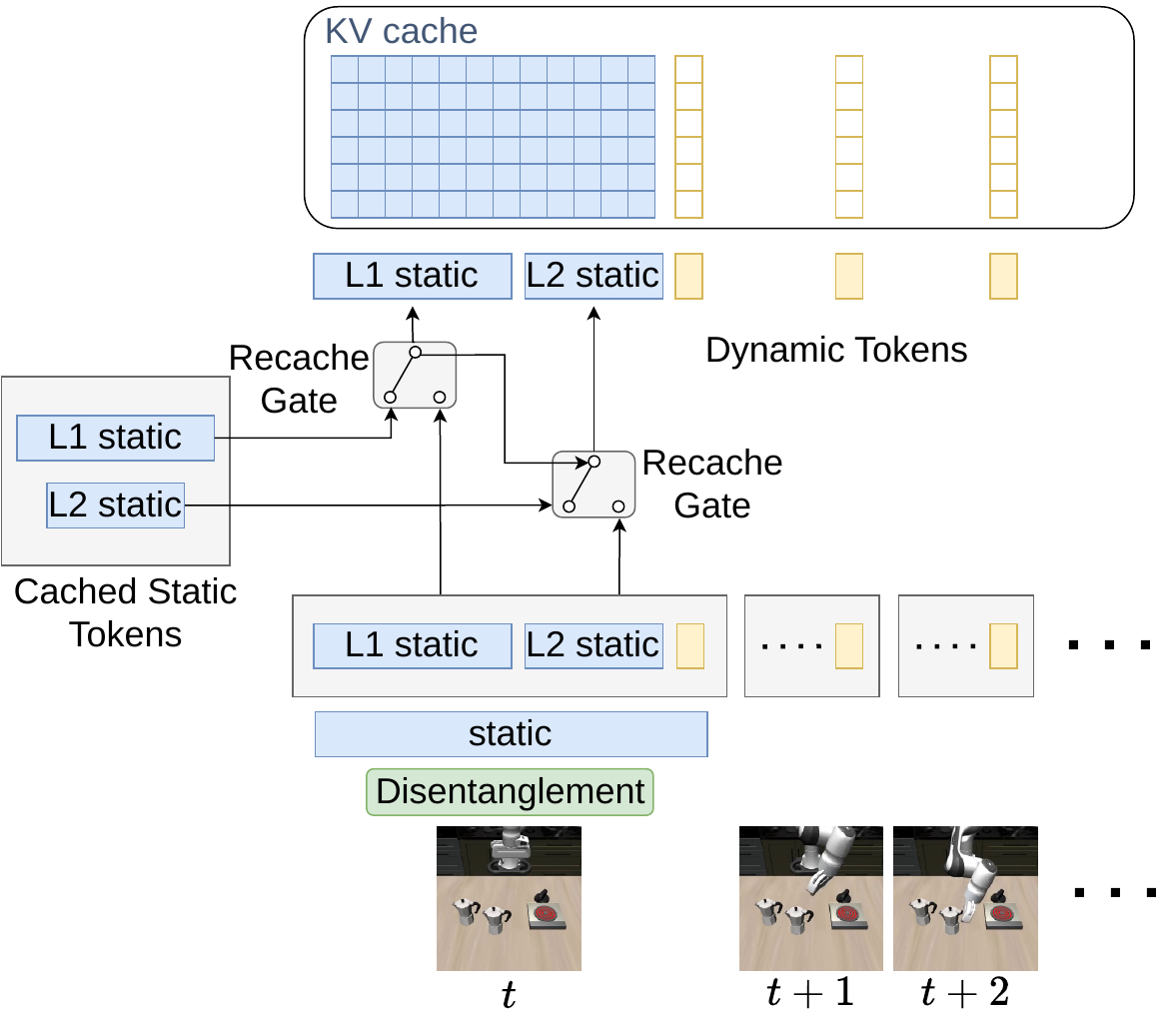}
    }
    \caption{Model architecture overview. We illustrate the design using two levels of static cache. At each level, a recache gate determines whether the cached static tokens should be reused or refreshed. If the L1 cache is refreshed, the L2 cache is also forcibly refreshed.}
    \label{fig:arch}
    % \vspace{-1em}
\end{wrapfigure}

where $\bs{a}_t$ is the action at the $t$ step. Most existing VLAs usually consider $T=0$ due to the limited context length of large language models.

\subsection{Model Architecture}
\noindent\textbf{Static-dynamic disentanglement}
Standard VLA vision backbones typically encode the image $\bs{X}_t$ into $N$ image tokens $\bs{Z}_t=\bs{z}_{t,1}, \bs{z}_{t,2}, \cdots, \bs{z}_{t,N}$ as inputs of the VLA's LLM backbone $\pi_\text{LLM}$. Such a design implicitly assumes that all visual tokens must be recomputed and reprocessed at every timestep. However, real-world environments exhibit strong temporal redundancy: many visual attributes remain unchanged across time, while only a subset varies dynamically. Moreover, static information exists at different temporal scales. For example, global scene layout or background structure may persist for long horizons, whereas object-level appearance may change more frequently due to occlusion or interaction. To reflect this structure, we disentangle visual tokens into multi-level static tokens and dynamic tokens.

\begin{equation}
\bs{Z}^{s_1}_t, \bs{Z}^{s_2}_t, \cdots, \bs{Z}^d_t = \\
  \underbrace{\bs{z}_{t,1}, \cdots, \bs{z}_{t,N_{s_1}}}_\text{static tokens (level 1)}, \underbrace{\bs{z}_{t,1}, \cdots, \bs{z}_{t,N_{s_2}}}_\text{static tokens (level 2)}, \cdots, \underbrace{\bs{z}_{t,1}, \cdots, \bs{z}_{t, N_d}}_\text{dynamic tokens}
\end{equation}

where $\bs{Z}^{s_l}_t$ denotes static tokens at level $l$, and $\bs{Z}^{d}_t$ are dynamic tokens. The ratio between static and dynamic tokens is set as a predefined hyperparameter. Under this design, dynamic tokens capture timestep-specific information and are recomputed at every timestep, whereas static tokens at each level are selectively reused through a learned caching mechanism, described later.

\noindent\textbf{Temporal dependency modeling}
Most VLAs \cite{kim2024openvla,li2024cogact,qu2025spatialvla} are only able to process the current observation ($T=0$),
\begin{equation}
    \bs{a}_t, \bs{a}_{t+1}, \cdots = \pi_\text{LLM}(\bs{Z}_t),
\end{equation}
where $\pi_\text{LLM}$ is the LLM backbone of the VLA.
With successful static-dynamic disentanglement, we only need to keep \emph{one copy} of the static tokens within a period and construct the input from multiple observations as

\begin{equation}
    \bs{a}_t, \bs{a}_{t+1}, \cdots = \pi_\text{LLM}(\bs{Z}^{s_1}, \bs{Z}^{s_2}, \cdots,\bs{Z}^d_{t-T}, \cdots, \bs{Z}^d_t),
\end{equation}

This formulation enables the model to leverage longer temporal context without duplicating static information, effectively alleviating context window bottlenecks while preserving relevant visual cues.

\noindent\textbf{Learning when to recache}
A key challenge is deciding when cached static tokens should be refreshed. Naively recomputing static tokens at every timestep negates any computational benefit, while overly aggressive reuse of static tokens risks stale representations. To address this, we introduce a learned recache gate at each static level $l$,

\begin{equation}
    g_l(\bs{Z}_{t-\Delta}, \bs{Z}_t) \in [0, 1],\quad \Delta=1,2,\cdots, 
\end{equation}

which predicts the probability that static tokens should be recomputed given the current observation and a cached reference from $\Delta$ timesteps earlier. During training, we use the Gumbel-softmax trick \cite{jang2016categorical,maddison2016concrete} to allow end-to-end differentiable binary decisions. At inference time, static tokens are refreshed if the probability is greater than a threshold, i.e., $g_l > \delta_l$; otherwise, the previous cache is reused. It is also worth noting that if the higher-level cache (e.g., L1) should be refreshed, the lower-level cache (e.g., L2) should also be refreshed during both training and inference. Additional details of the architecture of the recache gate could be found in Appendix~\ref{app:arch}.

\noindent\textbf{Computational and Acceleration Analysis}
Let $N$ be the total number of tokens per observation, $r$ the fraction of cached static tokens, and $T$ the number of observations in the context. Our static-dynamic disentanglement reduces the effective context length from $NT$ to $NT - rN(T-1)$, and reduces the FLOPs of the LLM backbone by a factor of $1-r$. The recache gate adds only negligible overhead ($\approx 1.27\%$). A full derivation and wall-clock latency breakdown are provided in Appendix~\ref{app:complexity}.

\begin{wrapfigure}[18]{r}{0.24\textwidth}
\vspace{-3em}
    \centering
    \noindent\makebox[\linewidth]{%
    \includegraphics[width=0.98\linewidth]{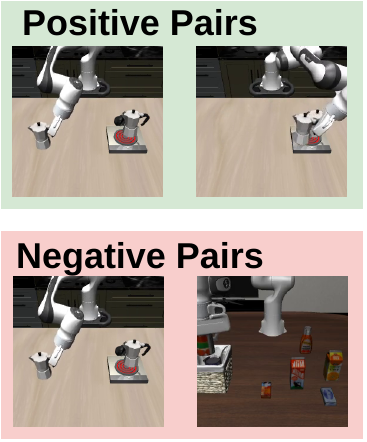}
    }
    \caption{Contrastive loss used to train static tokens to be temporally persistent. Observations from the same trajectory form positive pairs, while observations from different trajectories form negative pairs.}
    \label{fig:contrastive}
    % \vspace{-1em}
\end{wrapfigure}

\subsection{Training Objective}

Besides the standard task loss of the underlying VLA base model $\mathcal{L}_\text{task}$, we introduce two additional training objectives. The first promotes temporally persistent static tokens, and the second trains the cache gate that adaptively determines when cached representations should be refreshed.

\noindent\textbf{Learning static tokens} As illustrated in Figure~\ref{fig:contrastive}, to ensure that static tokens can be safely and effectively reused across time, they must remain stable over a temporal window while still encoding task-relevant information. To encourage this property, we apply a contrastive term to the static tokens. Specifically, observations from different timesteps within the same trajectory are treated as positive pairs, while observations from different trajectories are treated as negative pairs. We employ InfoNCE loss~\cite{oord2018representation} $\mathcal{L}_{\text{InfoNCE}}^{l}$ to perform contrastive learning for static tokens at level $l$.

% \begin{equation}
% \begin{aligned}
%     &\mathcal{L}_\text{InfoNCE}^l= \\
%     &\frac{1}{TN_{s_l}} \sum_{\Delta=1}^T \sum_{i=1}^{N_{s_l}} \frac{\exp(\bs{z}^{s_l}_{t,i} \cdot \bs{z}^{s_l}_{t-\Delta, i})}{\exp(\bs{z}^{s_l}_{t,i} \cdot \bs{z}^{s_l}_{t-\Delta, i}) + \sum_{z^-}\exp(\bs{z}^{s_l}_{t,i} \cdot \bs{z}_{}^-)},
% \end{aligned}
% \end{equation}
% where $N_{s_l}$ is the number of static tokens at level $l$. $\bs{z}^-$ are negative samples (i.e., observations from different trajectories). 

% Optimizing this objective encourages static tokens to remain consistent across timesteps within a trajectory, thereby making cache reuse safer and more effective during inference.

\noindent\textbf{Training the recache gate}
If the recache gate is trained only through task supervision, it tends to recompute static tokens at every timestep, which eliminates any computational benefit. To discourage this trivial solution, we add a regularization term that biases the gate toward reuse when observations are close in time.

\begin{equation}
    \mathcal{L}_\text{gate} = -p_\Delta \log g - (1-p_\Delta)\log (1-g),
\end{equation}

where $p_{\Delta} = 1 - e^{-\lambda \Delta}$ is a predefined prior which allows the use of current observation only when $\Delta$ is large. The recache gate learns to adaptively refresh static tokens only when necessary, balancing computational efficiency and model performance. Note that since $\lambda$ and the recache threshold $\delta_l$ jointly determine the recaching frequency, we fix $\lambda$ and expose only $\delta_l$ as the operating-point knob. 

The full training objective is
\begin{equation}
    \mathcal{L} = \mathcal{L}_\text{Task} + \sum_l \alpha_l  \mathcal{L}^l_\text{InfoNCE} + \beta \mathcal{L}_\text{gate},
\end{equation}
where $\alpha_l$ and $\beta$ weight the auxiliary regularization terms. A sensitivity analysis for these coefficients is provided in Appendix~\ref{app:sense}.

\subsection{LIBERO-Memory Benchmark}

Existing VLA simulation benchmarks \cite{liu2023libero, li2024evaluating} are designed for memoryless tasks, which do not require multi-frame integration, and the current observation alone is theoretically sufficient to predict the next action. For example, in a task with the goal “put X in the basket”, \cite{liu2023libero} the agent can act optimally without retaining information from previous frames. As a result, it is unclear whether the reported performance of existing models for modeling temporal dependency \cite{jang2025contextvla,shi2025memoryvlaperceptualcognitivememoryvisionlanguageaction} truly stems from improved temporal modeling or from other artifacts. This highlights the absence of a strong and fair benchmark for temporal dependency modeling and multi-frame integration.

To address this gap, we introduce a new simulation benchmark designed to test multi-frame integration explicitly. Our benchmark includes tasks that require a robot to retain and utilize information from past observations, thereby directly evaluating a model's temporal dependency modeling ability.

\begin{figure}[t]
    \centering
    \includegraphics[width=\linewidth]{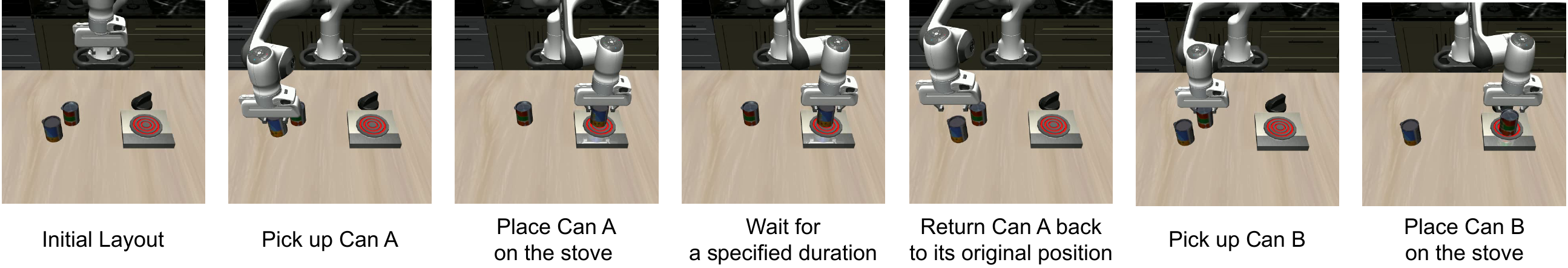}
    \caption{An example of the procedure of the proposed benchmark. }
    \label{fig:libero_stove}
    % \vspace{-1em}
\end{figure}

Following the setup of LIBERO, the robot operates in a tabletop environment containing three objects: two visually distinct cans and a stove. As illustrated in Figure~\ref{fig:libero_stove}, each episode consists of three tasks which reflects the structure of episodic memory \cite{tulving1972episodic,clayton1998episodic}.

\noindent\textbf{Tasks}
Each trial consists of three subtasks, each intentionally designed to require the model to retain and reason over information from previous observations at different aspects: 1) Grasp one of the cans, as specified by the instruction, and place it on the stove to heat. 2) After a period of time (given in the instructions), remove the can from the stove and return it to its original location. 3) Once the position of the first can has been restored, grasp the remaining can and place it on the stove.

This task design requires the model to retain \emph{episodic memory}, integrating information about what happened, where it occurred, and when it took place \cite{tulving1972episodic,clayton1998episodic}. First, the robot must remember the initial spatial layout in order to return the first can to its original position (where). Second, it must track the elapsed time since the can was placed on the stove to determine when to remove it (when). Third, it must remember which can has already been heated in order to correctly select the remaining one for the final step (what).

These requirements prevent the task from being solved using only the current observation or a short observation history. We use Robosuite \cite{zhu2025robosuitemodularsimulationframework} as the simulation framework to produce the dataset. The layout and the oracle demonstrations are programmatically generated. More details can be found in Appendix \ref{app:dataset_memory}.

\section{Experiments}
In this section, we evaluate the model’s inference efficiency as well as its multi-frame integration ability. We further present ablation studies of the key components of our method and analyze the model’s behavior through attention-map visualizations.

\subsection{Acceleration}
In this study, we validate our method's performance under acceleration by enabling KV caching of static components. 

\noindent\textbf{Baselines}
For baselines, we compare to those that also adopt temporal information reuse. \emph{FlashVLA} \cite{tan2025think} simply reuses the previous action based on a heuristic criterion.
\emph{TTF-VLA} \cite{liu2025ttf} and \emph{VLA-cache} \cite{xu2025vla} identify reusable visual patches using hand-designed heuristics and cache the corresponding embeddings from previous steps; however, these cached patches are not guaranteed to remain static in the hidden representation space. All methods are based on the same VLA base model (i.e., CogACT~\cite{li2024cogact}), with all other necessary variables controlled consistently.

\begin{table*}[t]
\newcolumntype{M}{>{\raggedleft\arraybackslash}p{0.1\linewidth}}
\newcommand{\cchead}[1]{\multicolumn{1}{c}{#1}}
    \centering\textbf{}
\caption{Results on the SimplerEnv benchmark. Task suites follow the setting of \citet{qu2025spatialvla}. All baselines and our method are applied to the same VLA base model (i.e., CogACT~\cite{li2024cogact}).}
\label{tab:simplerenv}
    \resizebox{\linewidth}{!}{
\begin{tabular}{lcccccc|ccc}
\toprule
\multirow{3}{*}{Method} & \multicolumn{3}{c}{Visual Matching} & \multicolumn{3}{c}{Variant Aggregation} & \multicolumn{3}{c}{Acceleration}  \\
\cmidrule(lr){2-4}\cmidrule(lr){5-7}\cmidrule(lr){8-10}
& \makecell{Pick\\Can} & \makecell{Move\\Near} & Drawer & \makecell{Pick\\Can}  & \makecell{Move\\Near} & Drawer  & FLOPs & \makecell{Latency\\(ms)} & Speedup \\
 \midrule
  % RT-2-X \cite{vuong2023open} & $78.7$ & $77.9$ & $25.0$ & $82.3$ & $79.2$ & $35.3$ & $-$ & $-$ & $-$\\
  % Octo-Base \cite{team2024octo} & $17.0$ & $4.2$ & $22.7$ & $0.6$ & $3.1$ & $1.1$ & $-$ & $-$ & $-$ \\
  % OpenVLA \cite{kim2024openvla}  & $18.0$ & $56.3$ & $63.0$ & $60.8$ & $67.7$ & $28.8$ & $-$ & $-$ & $-$ \\
  % SpatialVLA \cite{qu2025spatialvla} & $81.0$ & $69.6$ & $59.3$ & $89.5$ & $71.7$ & $36.2$ & $-$ & $-$ & $-$ \\
  % $\pi_0$ \cite{black2024pi0visionlanguageactionflowmodel} & $88.0$ & $80.3$ & $56.0$ & $-$ & $-$ & $-$ & $-$ & $-$ & $-$ \\
  % \midrule
 CogACT \cite{li2024cogact} & $91.3$ & $\underline{85.0}$ & $71.8$ &$89.6$ & $\underline{80.8}$ & $28.3$  & 100\% & $1360$ & $1.00\times$ \\
  + FlashVLA \cite{tan2025think} & $80.0$ & $57.1$ & $\underline{73.6}$ & $82.5$ & $60.8$ & $28.6$ & $83.4\%$ & $1024$ & $1.33\times$\\
 + TTF \cite{liu2025ttf} & $89.3$ & $71.3$ & $58.4$ & $88.0$ & $62.7$ & $\underline{34.0}$ & $86.5\%$ & $1051$ & $1.29\times$  \\
 + VLA-Cache \cite{xu2025vla} &  $\underline{92.0}$ & $83.3$  & $70.5$ & $\underline{91.7}$ & $79.3$  & $32.5$ & $80.1\%$ & $985$ & $1.38\times$ \\
 + \method{} (Ours) & $\mathbf{92.7}$ & $\mathbf{88.8}$ & $\mathbf{75.0}$ & $\mathbf{92.4}$  & $\mathbf{81.0}$ & $\mathbf{38.9}$ & $\mathbf{43.4\%}$ & $\mathbf{601}$ & $\mathbf{2.26\times}$ \\
 \bottomrule
\end{tabular}
}
\end{table*}

\begin{table}[t]
\centering
\caption{Results on the LIBERO benchmark. Policy inputs: third-person image, language instruction. All baselines and our method are applied to the same VLA base model (i.e., OpenVLA-OFT~\cite{kim2025fine}).}
\label{tab:libero}
\resizebox{\linewidth}{!}{
\begin{tabular}{lccccc|ccc}
\toprule
Models & Spatial & Object & Goal & Long & Average & FLOPs & Latency (ms) & Speedup \\
\midrule
% Diffusion Policy \cite{chi2025diffusion} & 78.3 & 92.5 & 68.3 & 50.5 & 72.4 & $-$ & $-$& $-$ \\
% Octo \cite{team2024octo} & 78.9 & 85.7 & 84.6 & 51.1 & 75.1 & $-$ & $-$& $-$\\
% DiT Policy \cite{hou2410diffusion} & 84.2 & 96.3 & 85.4 & 63.8 & 82.4 & $-$ & $-$& $-$\\
% OpenVLA \cite{kim2024openvla} & 84.7 & 88.4 & 79.2 & 53.7 & 76.5 & $-$ & $-$& $-$\\
% \midrule
OpenVLA-OFT \cite{kim2025fine} & $\underline{96.2}$ & $\mathbf{98.3}$ & $\mathbf{96.2}$ & $\underline{90.7}$ & $\underline{95.4}$ & $100.0\%$ & $741$ & $1.00\times$\\
% \midrule
+ FlashVLA \cite{tan2025think} & $71.8$ & $90.0$ &  $79.2$& $78.8$ & $80.0$ &  $88.7\%$& $657$ & $1.13\times$  \\
+ TTF \cite{liu2025ttf} & $95.8$ & $94.2$ &   $93.6$ &  $86.0$  & $92.4$ & $75.0\%$ & $575$ & $1.29\times$\\
% EfficientVLA & & & & 89.8 & \\
+ VLA-Cache \cite{xu2025vla} & $95.6$ & $92.8$ & $94.0$ & $89.6$ & $93.0$ & $85.5\%$ & $652$ & $1.14\times$ \\
+ \method{} (Ours) & $\mathbf{97.8}$ & $\underline{97.6}$ & $\mathbf{96.2}$ & $\mathbf{92.4}$ & $\mathbf{96.0}$ & $\mathbf{63.4}\%$ & $\mathbf{437}$ & $\mathbf{1.70}\times$ \\
\bottomrule
\end{tabular}
}
\end{table}

\noindent\textbf{Simulation benchmark}
We use SimplerEnv~\cite{li2024evaluating} and LIBERO~\cite{liu2023libero} task suite to evaluate the performance and acceleration. \citet{kim2025fine} offers two settings for LIBERO, one using only a third-person image and a language instruction as inputs, and another that additionally includes an extra camera view and proprioceptive state. We adopt the former setting, as it offers broader applicability and aligns more closely with standard VLA deployment scenarios. For the LIBERO benchmark, we finetune the model to the corresponding downstream datasets. For the SimplerEnv benchmark, we use Open X-Embodiment (OXE) \cite{vuong2023open} as the training dataset. All settings are consistent with the corresponding original base model. Dataset statistics could be found in Appendix \ref{app:dataset}.

\noindent\textbf{Real-World benchmark} We evaluate \method{} with Cobot Magic robots on 7 tasks across 3 categories, each with 20 trials. This evaluation scale is consistent with prior work~\cite{shi2025memoryvlaperceptualcognitivememoryvisionlanguageaction,li2024cogact}. Models are finetuned on each task. More details could be found in Appendix~\ref{app:real}.

\noindent\textbf{Evaluation} The performance is measured by success rate, and acceleration is measured by the FLOPs reduction rate and the inference latency.

\noindent\textbf{Implementation details} We use two levels of static cache. L1 static cache consists of 133 ($52\%$) tokens and L2 static cache consists of 107 ($42\%$) tokens, resulting in 16 dynamic tokens. The sensitivity analysis of the static ratio is in Appendix~\ref{app:static_ratio}. FLOPs are measured by open-source tools \footnote{\url{https://github.com/MrYxJ/calculate-flops.pytorch}}. Additional details, including the recaching threshold $\delta_l$, the resulting average recaching intervals and the latency distribution when rollout, can be found in Appendix~\ref{app:impl} and Appendix~\ref{app:lat_dist}.

\noindent\textbf{Results}
Table~\ref{tab:simplerenv}  and Table~\ref{tab:libero} summarize the performance and the acceleration results. Table~\ref{tab:real_res} summarizes the results on the real-world benchmark. Across tables, we show that our method achieves improved or comparable performance across benchmarks with substantial speedup. Notably, on the SimplerEnv benchmark, our method improves the base model by $4.9\%$, and outperforms the best baseline by $3.9\%$ and achieves $2.26\times$ acceleration.  On the real-world benchmark, our model improves the base model by $10.6\%$ and achieves $2.21\times$ acceleration.

\subsection{Multi-Frame Integration for Temporal-dependency Modeling} 

\noindent\textbf{Baselines}
We consider the following baselines: \emph{OpenVLA-OFT$^\dagger$} incorporates full image tokens from historical observations. Since this baseline incurs a prohibitively large context length compared with standard models, we allocate the maximum available resources and tune it for the best achievable performance. \emph{TTF-VLA} \cite{liu2025ttf} mixes the current observation with past observations in a patch-wise manner. \emph{TraceVLA} \cite{zheng2024tracevla} visualizes the trajectories of active points by overlaying them onto the image using distinct colors, referred to as visual traces. \emph{MemoryVLA} \cite{shi2025memoryvlaperceptualcognitivememoryvisionlanguageaction} feeds the LLM backbone with images from each timestep independently and extracts a representation for each; only the multi-timestep representations are provided to the decoder. \emph{ContextVLA} \cite{jang2025contextvla} pools historical observations into a fixed number of tokens and prepends them to the image tokens of the current observation. All methods are applied to the same VLA base model (i.e., OpenVLA-OFT~\cite{kim2025fine}), with all other necessary variables controlled consistently.

\begin{table}[t]
\centering
\caption{Success rate (\%) on real-world experiments with the Cobot Magic robot. All baselines and our method are applied to the same VLA base model (i.e., OpenVLA-OFT~\cite{kim2025fine}).}
\label{tab:real_res}
\resizebox{\linewidth}{!}{
\begin{tabular}{l|cccc|cc|c|ccc}
\toprule
\multirow{3}{*}{Method}
& \multicolumn{4}{c|}{Pick-and-Place}
& \multicolumn{2}{c|}{Pouring}
& \multicolumn{1}{c|}{Container}
& \multicolumn{3}{c}{Acceleration Metrics} \\
\cmidrule(lr){2-5}
\cmidrule(lr){6-7}
\cmidrule(lr){8-8}
\cmidrule(lr){9-11}

& \makecell{Corn\\$\rightarrow$Pot}
& \makecell{Corn\\$\rightarrow$Plate}
& \makecell{Bread\\$\rightarrow$Pan}
& \makecell{Carrot\\Transfer}
& \makecell{Pour\\Coke}
& \makecell{Pour\\Bottle}
& \makecell{Open\\Drawer}
& FLOPs
& \makecell{Latency\\(ms)}
& Speedup \\
\midrule

OpenVLA-OFT
& $55$ & $\underline{70}$ & $\mathbf{55}$ & $\underline{65}$
& $\underline{60}$ & $35$
& $\mathbf{35}$
& $100.0\%$
& $1089$
& $1.00\times$ \\

+ FlashVLA
& $35$ & $55$ & $30$ & $50$
& $55$ & $30$
& $15$
& $\underline{62.5\%}$
& $\underline{791}$
& $\underline{1.30\times}$ \\

+ TTF
& $35$ & $\underline{70}$ & $30$ & $55$
& $55$ & $\underline{40}$
& $15$
& $98.6\%$
& $1077$
& $1.01\times$ \\

+ VLA-Cache
& $\underline{60}$ & $65$ & $35$ & $50$
& $45$ & $\underline{40}$
& $30$
& $74.0\%$
& $836$
& $1.27\times$ \\

% \midrule
+ \method{} (Ours)
& $\mathbf{65}$ & $\mathbf{80}$ & $\underline{50}$ & $\mathbf{70}$
& $\mathbf{60}$ & $\mathbf{50}$
& $\mathbf{35}$
& $\mathbf{30.5\%}$
& $\mathbf{493}$
& $\mathbf{2.21\times}$ \\
\bottomrule
\end{tabular}
}
\vspace{-2em}
\end{table}

\noindent\textbf{Simulation benchmark and evaluation} We adopt the proposed LIBERO-memory benchmark to evaluate multi-frame integration by memory-dependent tasks, whose designs are inspired by the notion of episodic memory. The benchmark comprises three objectives.
(1) \emph{(where) Position Reset.} After heating the first can, the robot is required to remember its original position and return it to the position. Performance is measured by success rate, where an episode is considered successful if the positional error is below a threshold.
(2) \emph{(when) Doneness.} The model must track the elapsed heating time and remove the can at the appropriate moment. Given a target heating duration specified in the instruction, we measure the absolute difference (in seconds) between the actual and desired heating times, capturing overcooking or undercooking.
(3)\emph{(what) On-Stove.} After heating and resetting the first can, the robot is required to remember which can has been cooked and place the second can on the stove for heating. We report the success rate of placing the second can on the stove.

\begin{wraptable}[12]{r}{0.65\textwidth}
\vspace{-1em}
    \centering
    \caption{Results on the temporal-dependent tasks for multi-frame integration. All models use OpenVLA-OFT \cite{kim2025fine} as the base model. OpenVLA-OFT$^\dagger$ denotes a naive baseline that packs as many frames as possible into the context.}
    \label{tab:stove_res}
    \resizebox{\linewidth}{!}{
    \begin{tabular}{lccc}
    \toprule
        Models & On Stove $\uparrow$ & Position Reset $\uparrow$ & Doneness (seconds) $\downarrow$ \\
    \midrule
        OpenVLA-OFT$^\dagger$ & $39.0\%$ & $78.0\%$ & $0.44$ \\
        TTF-VLA \cite{liu2025ttf} & $7.8\%$ & $1.5\%$ & $1.51$ \\
        TraceVLA \cite{zheng2024tracevla} & $2.0\%$ & $3.0\%$ & $1.41$ \\
        MemoryVLA \cite{shi2025memoryvlaperceptualcognitivememoryvisionlanguageaction} & $23.0\%$ & $2.0\%$ & $1.49$ \\
        ContextVLA \cite{jang2025contextvla} & $50.8\%$ & $22.3\%$ & $0.37$ \\
        \method{} (Ours) & $\mathbf{69.8}\%$ & $\mathbf{83.0}\%$ &  $\mathbf{0.26}$ \\
    \bottomrule
    \end{tabular}
    }
\end{wraptable}

\noindent\textbf{Real-world benchmark}
We evaluate DySta with Cobot Magic robots on 3 tasks, each with 20 trials. Models are finetuned on each task. Performance is measured by success rate. More details could be found in Appendix \ref{app:real}.

\noindent\textbf{Implementation details} We use one static level. The ratio of static tokens is set to $0.9$, corresponding to $230$ static tokens and $26$ dynamic tokens. The number of observations is set to $20$, with a sampling interval of $20$ frames (1 frame per second), resulting in $750$ context tokens. More details could be found in Appendix~\ref{app:impl}.

\noindent\textbf{Results} 
As shown in Table~\ref{tab:stove_res}, our method significantly and consistently outperforms all baselines on all objectives, with an average improvement of $24.5\%$. On the real-world experiments in Appendix~\ref{app:real_mem}, we achieve $23.3\%$ absolute in absolute success rate.
Single-image methods such as TraceVLA almost always fail, since the task cannot be solved from the current observation alone, and visual traces provide only marginal additional benefit. 
ContextVLA underperforms due to non-learnable pooling observations, which may discard critical temporal and spatial details.
MemoryVLA presents the LLM backbone with only a single frame at each timestep, leaving multi-frame reasoning to a lightweight decoder; as a result, the LLM itself does not reason over multiple frames, which limits the model's capacity.
While these baselines have their own ways to reduce context length, they do so at the cost of information or model capacity. In contrast, our approach preserves all relevant visual information by reusing temporally persistent static tokens, enabling effective multi-frame integration within a limited context window.

\subsection{Ablation Studies}

To validate the effectiveness of the proposed components, we conduct ablation studies on each of them, with results summarized in Table~\ref{tab:ablation}. From the second row, removing the contrastive learning objective leads to a noticeable performance degradation, as there is no longer an explicit mechanism to enforce temporal consistency in the static tokens. Performance also declines when the L2 static cache is removed and only a single-level static cache is used, as shown in the third row, highlighting the importance of the proposed multi-level caching design. Finally, replacing the learnable recache gate results in a further drop in performance, demonstrating the necessity of adaptive cache refreshing.

\begin{wraptable}[13]{r}{0.65\textwidth}
\vspace{-2em}
    \centering
        \caption{Ablation studies. \emph{w/o Contrast} removes the contrastive learning objective during training. \emph{w/o L2 cache} ablates the multi-level design by retaining only a single static cache level. \emph{Fixed step} replaces the learnable recache gate by forcing the refresh of the cache at fixed intervals. The refresh interval is set to match the average recaching interval of \method{} (first row).}
    \label{tab:ablation}
    \resizebox{\linewidth}{!}{
    \begin{tabular}{cccccc}
    \toprule
         & \multicolumn{2}{c}{visual matching} && \multicolumn{2}{c}{variant aggregation}  \\
         \cmidrule{2-3}
         \cmidrule{5-6}
         & PickCan & MoveNear & & PickCan & MoveNear \\
         \midrule
        \method{} & $\mathbf{92.7}$ & $\mathbf{88.8}$ & &$\mathbf{92.4}$ & ${81.0}$ \\ 
        w/o Contrast & $91.3$ & $84.6$ & &$91.6$ & $\mathbf{82.7}$ \\
        w/o L2 cache & $92.0$ & $85.0$ & & $91.5$ & $81.5$ \\
        fixed step & $87.7$ & $81.7$ & & $85.3$ & $74.8$ \\
    \bottomrule
    \end{tabular}
    }
\end{wraptable}

\subsection{Visualizations}
\begin{wrapfigure}[22]{r}{0.5\textwidth}
\vspace{-12pt}
    \centering
\noindent\makebox[\linewidth]{    \includegraphics[width=1.13\linewidth]{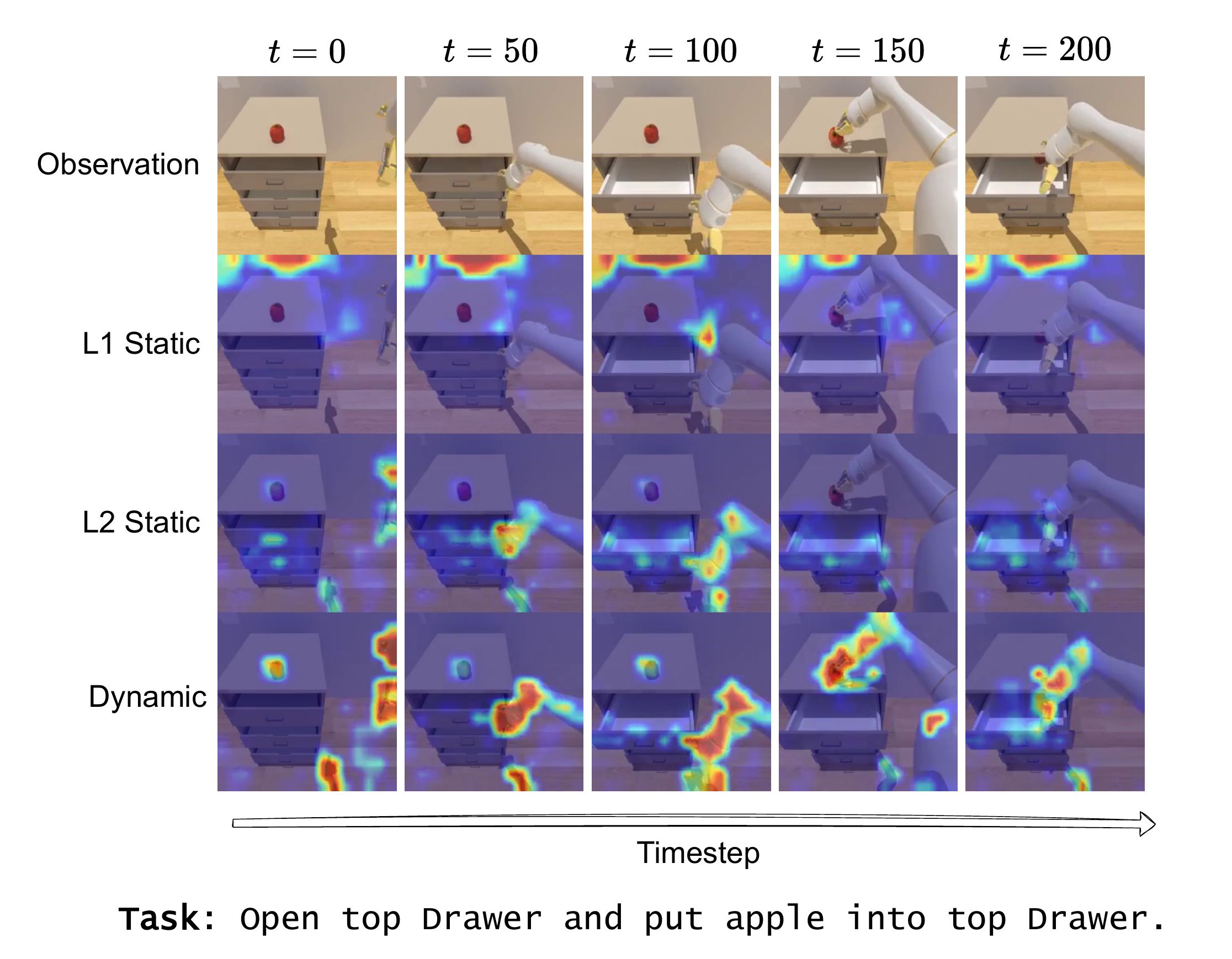} }
% \vspace{-15pt}
    \caption{Attention map visualization across time. For each token, we compute its last-layer attention to image patches and upsample the result to the full image resolution to produce a heatmap. Heatmaps are averaged over tokens of the same type and displayed by row (e.g., the Dynamic row shows the average attention heatmap of all dynamic tokens).}
    \label{fig:attention}
\end{wrapfigure}

We visualize the attention between the image and static/dynamic tokens across timesteps in Figure~\ref{fig:attention}.

Dynamic tokens consistently attend to movable objects, most notably the gripper (including its shadow) and the apple on the tabletop. This behavior aligns with their intended role of capturing temporally varying, action-relevant elements in the scene.

L1 static tokens, which are designed to represent the most persistent visual information, primarily focus on background and ambient regions. Their attention heatmaps are often strongest in areas without salient foreground objects. We interpret this behavior as L1 static tokens of shallow layers functioning as sink tokens, capturing coarse, global scene context rather than object-specific details. Importantly, their attention patterns remain highly consistent across timesteps, reflecting strong temporal invariance.

L2 static tokens exhibit intermediate behavior between L1 static and dynamic tokens. They attend more strongly to semi-static objects, such as the drawer, which typically remains stationary most of the time. At $t=0$ both drawer handles are highlighted, as either could potentially be moved. After the drawer is opened ($t>100$), the attention maps stabilize and remain focused on the drawer structure, indicating that these tokens capture object-level static information with moderate temporal persistence. The robotic arm is also highlighted at certain frames, which we attribute to its importance in the task and the need to preserve its appearance across timesteps.

We further illustrate how static-dynamic assignments evolve along a trajectory in Appendix~\ref{app:time_evo}.

\section{Conclusion}
We introduced \method{}, a method that improves efficiency and multi-frame integration in VLA models via static–dynamic disentanglement. By reusing temporally persistent static tokens and selectively refreshing them through a learnable recache gate, \method{} enables compact multi-frame contexts and efficient inference. Experiments show that our approach substantially improves performance on temporally dependent tasks while providing meaningful speedups over strong baselines. These results suggest that explicitly modeling temporal persistence is a promising direction for scalable and practical VLA systems. \noindent\textbf{Limitations}. We realize our approach based on pretrained VLAs. Such a strategy may not fully unleash the performance of the model. Future work could focus on pretraining VLAs with our architecture from scratch.

\newpage
\bibliography{reference}
\bibliographystyle{plainnat}
\newpage
\appendix
\section{Dataset and Benchmark Details}
\label{app:dataset}

\subsection{Simulation}
\paragraph{Training}
Table~\ref{tab:mixture} summarizes the Open-X-Embodiment dataset we use for the base model CogACT \cite{li2024cogact}. Table~\ref{tab:libero-data} summarizes the LIBERO dataset we use for the base model OpenVLA-OFT~\cite{kim2025fine}.

\begin{table}[htbp]
\centering
    \caption{Open-X-Embodiment dataset composition we use for CogACT training.}
    \label{tab:mixture}
    \begin{tabular}{lr}
    \toprule
     Dataset & Ratio \\
    \midrule
     Fractal~\citep{brohan2022rt} & 27.1\% \\
     Kuka~\citep{kalashnikov2018scalable} & 14.7\% \\
     Bridge~\citep{walke2023bridgedata,Ebert2021BridgeDB} & 15.3\% \\
     Taco Play~\citep{rosete2023latent,mees2023grounding} & 3.4\% \\
     Jaco Play~\citep{dass2023jacoplay} & 0.6\% \\
     Berkeley Cable Routing~\citep{luo2024multi} & 0.3\% \\
     Roboturk~\citep{mandlekar2018roboturk} & 2.7\% \\
     Viola~\citep{zhu2023viola} & 1.1\% \\
     Berkeley Autolab UR5~\citep{BerkeleyUR5Website} & 1.4\% \\
     Toto~\citep{zhou2023train} & 2.3\% \\
     Stanford Hydra Dataset~\citep{belkhale2023hydra} & 5.1\% \\
     Austin Buds Dataset~\citep{zhu2022bottom} & 0.2\% \\
     NYU Franka Play Dataset~\citep{cuiplay} & 1.0\% \\
     UCSD Kitchen Dataset~\citep{ucsdkitchens} & $<$0.1\% \\
     Austin Sailor Dataset~\citep{nasiriany2023learning} & 2.5\% \\
     Austin Sirius Dataset~\citep{liu2022robot} & 2.0\% \\
     DLR EDAN Shared Control~\citep{quere2020shared} & $<$0.1\% \\
     IAMLab CMU Pickup Insert~\citep{saxena2023multi} & 1.0\% \\
     UTAustin Mutex~\citep{shahmutex} & 2.6\% \\
     Berkeley Fanuc Manipulation~\citep{zhu2023fanuc} & 0.9\% \\
     CMU Stretch~\citep{mendoncastructured} & 0.2\% \\
     BC-Z~\citep{jang2022bc} & 8.6\% \\
     FMB Dataset~\citep{luo2023fmb} & 2.4\% \\
    \bottomrule
    \end{tabular}
\end{table}

\begin{table}[htbp]
    \caption{LIBERO dataset}
    \label{tab:libero-data}
    \centering
    \begin{tabular}{c|cccc}
    \toprule
         Dataset & Libero-Spatial & Libero-Object & Libero-Goal & Libero-Long \\
    \midrule
       \#(trajectories) & 432 & 454 & 428 & 379 \\
    \bottomrule
    \end{tabular}

\end{table}

\paragraph{SimplerEnv} SimplerEnv \cite{li2024evaluating} is a simulation-based benchmark designed for tabletop robotic manipulation. It is explicitly constructed to minimize the sim-to-real gap, demonstrating strong alignment between simulation performance and real-world execution across multiple robot platforms. The benchmark supports two complementary evaluation configurations: Visual Matching, which emphasizes high visual fidelity to real-world environments, and Variant Aggregations, which systematically introduce visual perturbations such as changes in background, lighting, distractors, and surface textures to assess robustness. SIMPLER includes tasks instantiated on both the Google Robot and the WidowX robot, covering a diverse set of manipulation primitives such as picking, placing, navigation, and articulated object interaction. In all tasks, agents receive RGB visual observations and natural language instructions. SIMPLER offers two evaluation settings:
\begin{itemize}[noitemsep,topsep=0pt,leftmargin=*]
    \item \emph{Visual Matching:} Real-world images are overlaid onto simulated environments, with foreground objects and robots adjusted to closely match real-world appearances.
    \item \emph{Variant Aggregation:} Multiple environmental variations are generated—such as different backgrounds, lighting conditions, and surface textures.
\end{itemize}

\paragraph{LIBERO} LIBERO is built around a simulated Franka Emika Panda arm and provides high-quality human-teleoperated demonstration data paired with language-conditioned tasks to support sample-efficient learning and generalization. The benchmark comprises four distinct task suites—LIBERO-Spatial, LIBERO-Object, LIBERO-Goal, and LIBERO-Long, each providing 10 distinct tasks.

\subsection{Real-World}
\label{app:real}

\begin{figure}
    \centering
    \includegraphics[width=\linewidth]{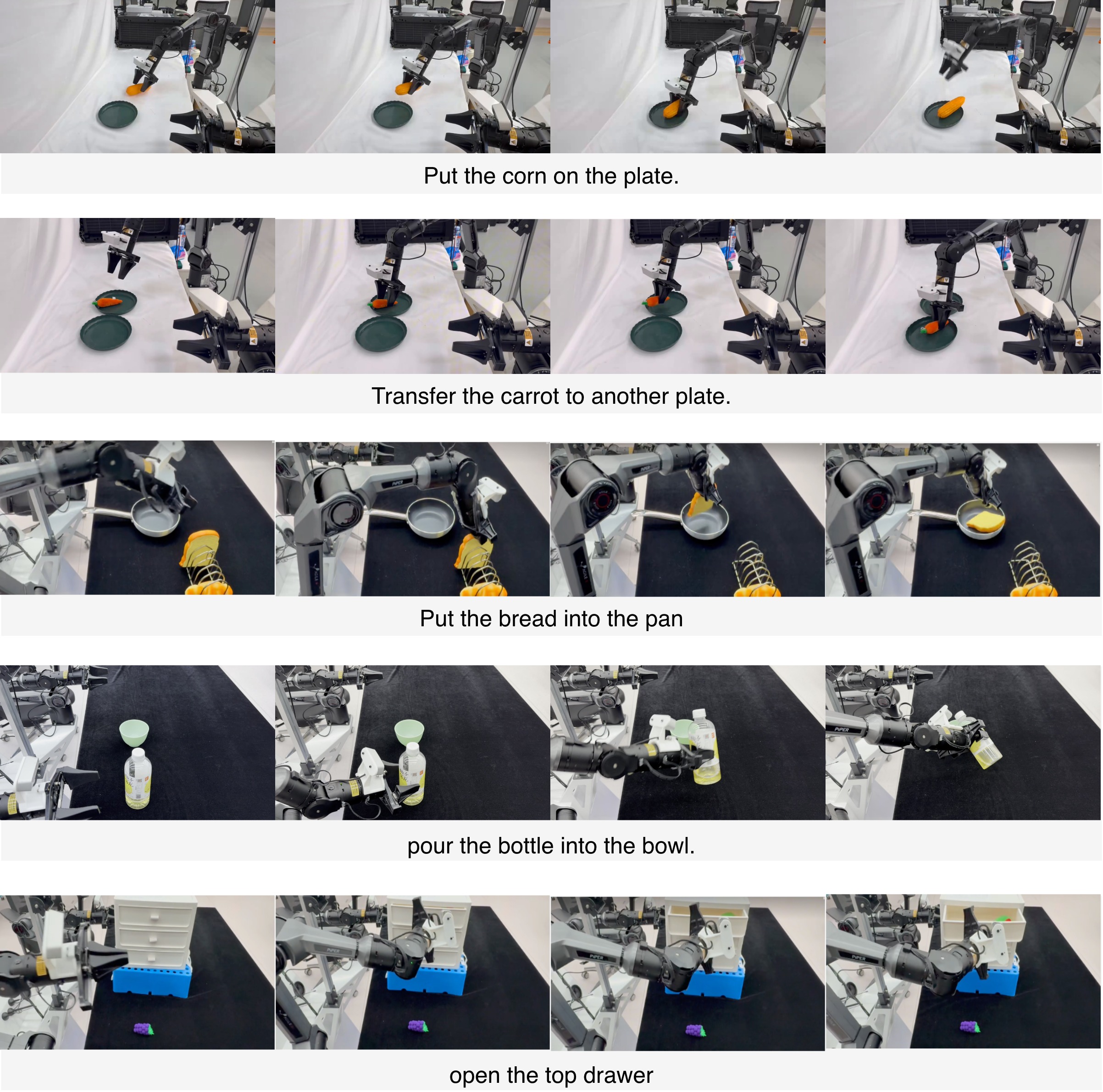}
    \caption{In real-world evaluations with the Cobot Magic robot, each task is evaluated over 20 episodes.}
    \label{fig:real}
\end{figure}
For the acceleration experiments, we include 7 tasks across 3 categories in the acceleration experiment, with 20 episodes used for training and 20 trials used for evaluating each task. The task designs are detailed below:
\begin{itemize}
\item \textbf{Pick-and-Place.} The robot is instructed to move an object from its original location to a specified destination (e.g., a pan or pot). We report the success rate for each object.
\item \textbf{Pouring.} The robot is instructed to pour the contents of a bottle or can into a container. We report the success rate for each target container.
\item \textbf{Drawer Opening.} The robot is instructed to open a specified drawer. We report the success rate of successfully opening the drawer.
\end{itemize}

For the multi-frame integration experiments, we evaluate three tasks that require memory or temporal reasoning:
\begin{itemize}
\item \textbf{Sequential Heat.} The robot is instructed to heat the vegetables one at a time. The model must remember which vegetables have already been heated.
\item \textbf{Pour Twice.} The robot is instructed to pour from the bottle twice. The model must keep track of whether the first pour has already been completed.
\item \textbf{Press Buttons.} The robot is instructed to press a sequence of buttons. The model must remember which buttons have already been pressed and determine the next button to press.
\end{itemize}

For each task, we vary both the object's location and the robotic arm's starting position. The demonstrations of tasks are illustrated in Figure~\ref{fig:real} and Figure~\ref{fig:memory_examples}. 

\begin{figure}
    \centering
    \includegraphics[width=\linewidth]{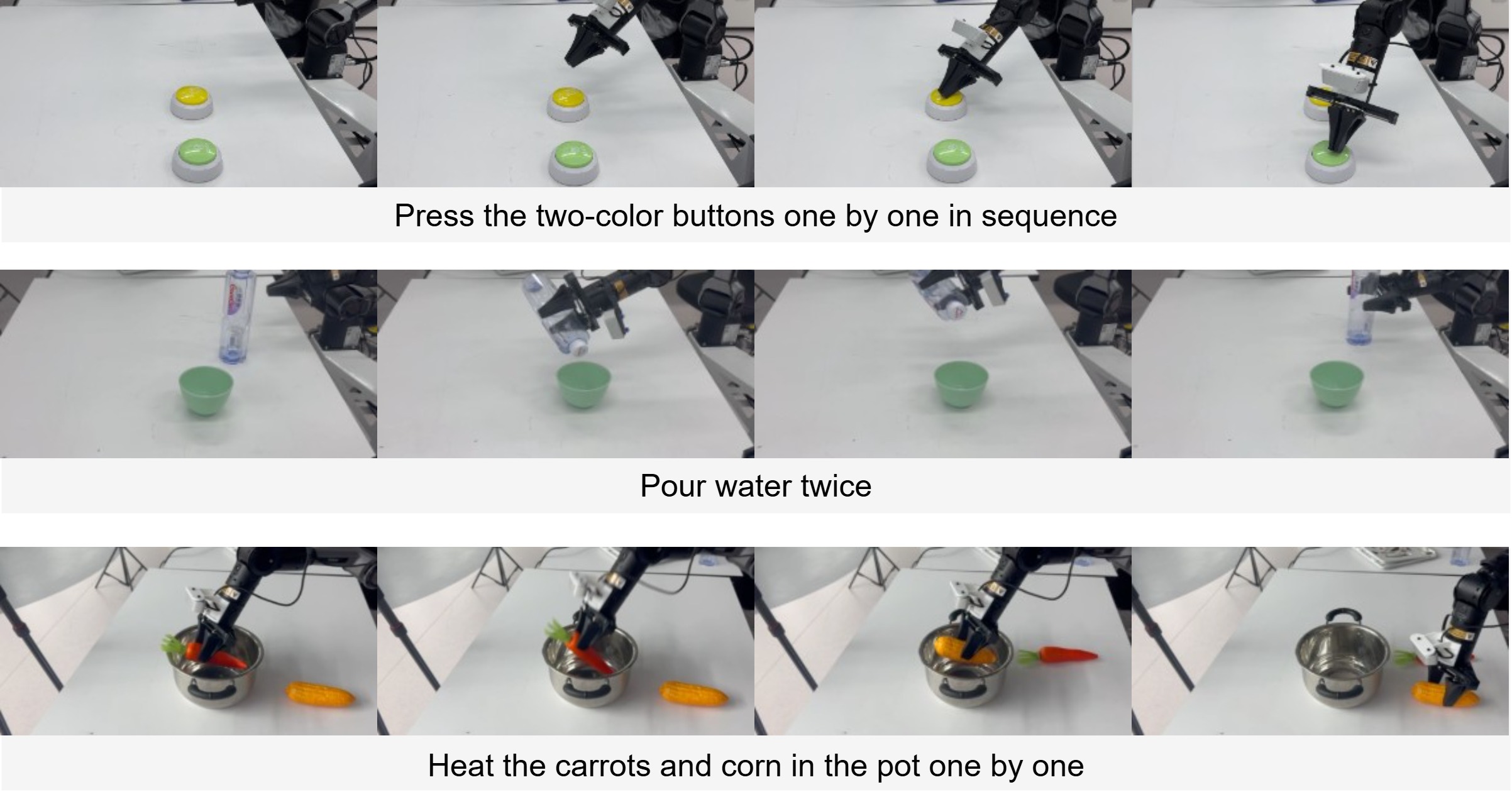}
    \caption{In real-world evaluations with the Cobot Magic robot for multi-frame integration, each task is evaluated over 20 episodes.}
    \label{fig:real}
\end{figure}

\section{Architecture Details}
\label{app:arch}

\paragraph{Recache gate}
The recache gate is designed to predict the probability of refreshing cached representations based on images (or their latent representations) observed at different timesteps. To minimize additional inference latency in the VLA, the recache gate must remain lightweight. Accordingly, we adopt a simple MLP-based architecture.

\begin{figure}
    \centering
    \includegraphics[width=\linewidth]{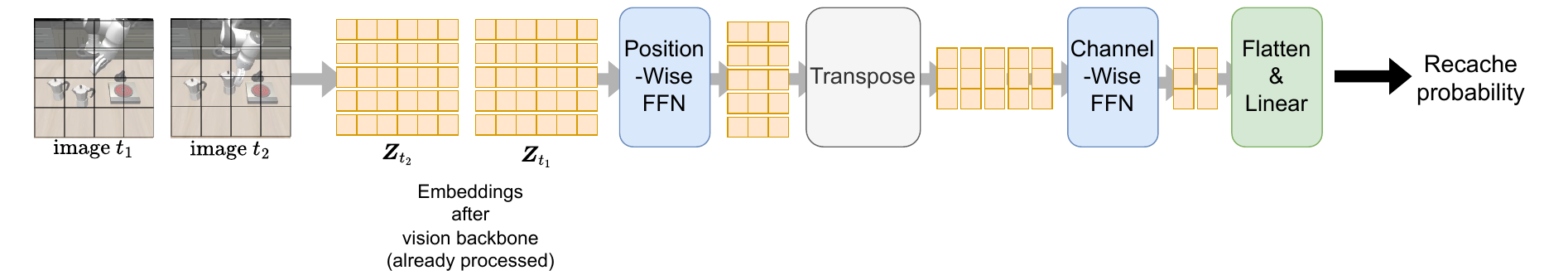}
    \caption{Architecture of the recache gate.}
    \label{fig:cache_gate}
\end{figure}

As illustrated in Figure~\ref{fig:cache_gate}, the gate takes as input latent embeddings $\bs{Z}_{t_1}$ and $\bs{Z}_{t_2}$ produced by the VLA’s vision backbone at two different timesteps. We first apply a position-wise feedforward network to each embedding. The embeddings are then transposed to enable a channel-wise feedforward operation. After these transformations and dimensionality reduction, the resulting features are flattened and passed through a final MLP to predict the recache probability. For different static levels, we share all parameters except for the final prediction head.

\section{LIBERO-Memory Benchmark}
\label{app:dataset_memory}
Following the LIBERO benchmark \cite{liu2023libero}, the layout initialization is specified by the BDDL scene description file, which is a file format that specifies the initial layout's sampling strategy and other meta information.

\begin{lstlisting}[language=PDDL, caption=An example of the PDDL scene description file used in LIBERO-Memory.]
(define (problem LIBERO_Kitchen_Tabletop_Manipulation)
  (:domain robosuite)
  (:language Heat the tomato sauce on the stove for 1.4 second and then heat the alphabet soup)
    (:regions
      (operation_region
          (:target kitchen_table)
          (:ranges (
              (-0.2 0.09999999999999999 0.0 0.2)
            )
          )
          (:yaw_rotation (
              (0.0 0.0)
            )
          )
      )
      (stove_region
          (:target kitchen_table)
          (:ranges (
              (-0.30000000000000004 -0.2 -0.1 -0.09999999999999999)
            )
          )
          (:yaw_rotation (
              (0.0 0.0)
            )
          )
      )
      (cook_region
          (:target flat_stove_1)
      )
    )

  (:fixtures
    kitchen_table - kitchen_table
    flat_stove_1 - flat_stove
  )

  (:objects
    alphabet_soup_1 - alphabet_soup
    tomato_sauce_1 - tomato_sauce
  )

  (:obj_of_interest
    tomato_sauce_1
    alphabet_soup_1
  )

  (:init
    (On tomato_sauce_1 kitchen_table_operation_region)
    (On alphabet_soup_1 kitchen_table_operation_region)
    (On flat_stove_1 kitchen_table_stove_region)
    (Turnon flat_stove_1)
  )

  (:goal
    (And (CloseXY tomato_sauce_1) (On alphabet_soup_1 flat_stove_1_cook_region))
  )

)
\end{lstlisting}

\paragraph{Oracle demonstrations}
We generate expert trajectories programmatically because the correct behavior in LIBERO-Memory is fully specified and unambiguous. For example, for a sub-task such as “place the tomato sauce can on the stove”, an oracle with access to the simulator state can deterministically execute the correct sequence. The oracle knows the target object and its current location, and issues movement commands that drive the end-effector toward that location. After each movement, it compares the updated end-effector position with the desired target and decides the next control direction. Once the end-effector is sufficiently close, the oracle closes the gripper to grasp the object and subsequently moves it to the specified target location.

\begin{figure}
    \centering
    \includegraphics[width=\linewidth]{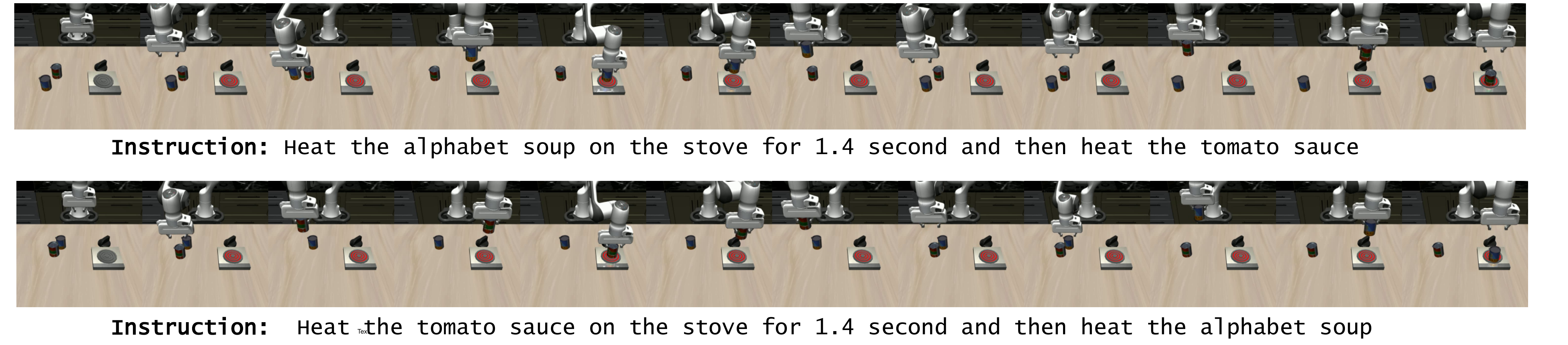}
    \caption{Examples of our proposed LIBERO Memory Benchmark.}
    \label{fig:memory_examples}
\end{figure}

\begin{figure}
    \centering
    \includegraphics[width=0.5\linewidth]{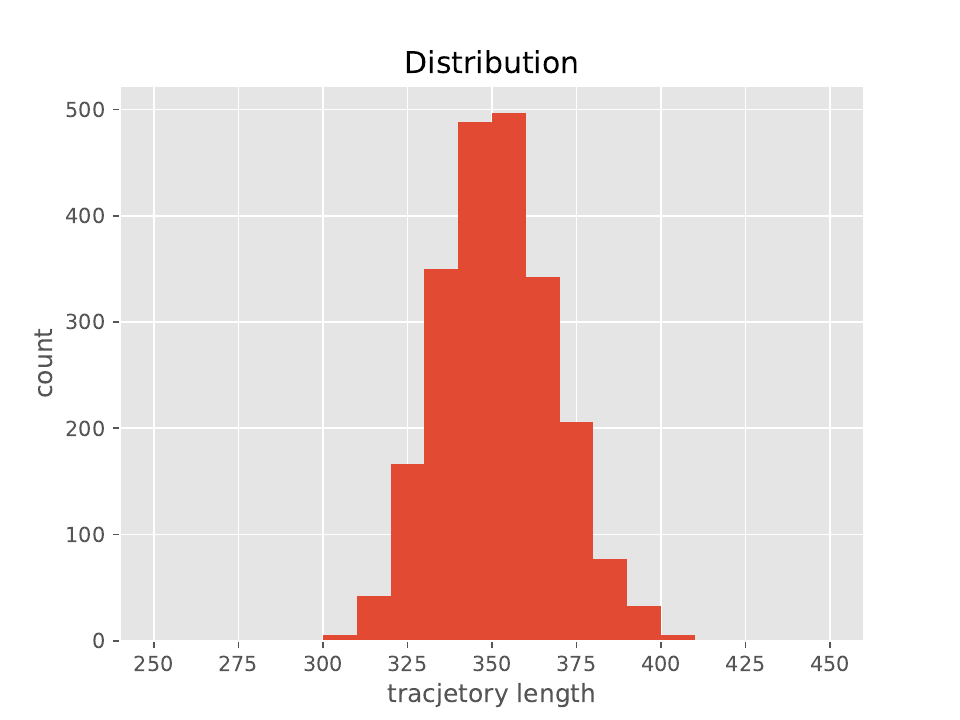}
    \caption{Distribution of the trajectory length of the LIBERO-Memory benchmark}
    \label{fig:stats}
\end{figure}

Figure~\ref{fig:memory_examples} shows more detailed example trajectories of the benchmark. Figure~\ref{fig:stats} shows the distribution of the trajectory length of the benchmark.

 To better contextualize the absolute performance of our model, we include a human baseline for the benchmark in Table~\ref{tab:stove_human}. Specifically, we recruited two human participants familiar with the task setup, who performed the tasks using standard keyboard controls. Each human participant completed 50 rollout episodes. The performance varies across subtasks: \emph{On Stove} is largely semantic and can be solved reliably by humans, while \emph{Position Reset} and \emph{Doneness} require precise tracking of spatial locations and elapsed time over long horizons, making them more challenging for humans and highlighting greater potential for machine learning models to excel.

\begin{table}[H]
    \centering
    \caption{Human Performance on LIBERO-Memory.}
    \label{tab:stove_human}
    % \resizebox{\linewidth}{!}{
    \begin{tabular}{lccc}
    \toprule
        Models & On Stove $\uparrow$ & Position Reset $\uparrow$ & Doneness $\downarrow$ \\
    \midrule
    Human 1 & $98.0\%$  & $74.0\%$ & $1.58$ \\
        Human 2 & $96.0\%$ & $68.0\%$ & $1.44$\\
        \method{} (Ours) & $69.8\%$ & $83.0\%$ &  $0.26$ \\
    \bottomrule
    \end{tabular}
    % }
\end{table}

\section{Implementation Details and Hyperparameter Choices}
\label{app:impl}
For the recache gate in Figure~\ref{fig:cache_gate}, the first FFN is a $3$-layer MLP that reduce the embedding dimension from $2\times4096$ to $128$. The second FFN is a $3$-layer MLP that reduce the number of embeddings from $256$ to $128$. The head after the flatten operation is a linear layer with input dimension $128 \times 64$.
 
\paragraph{CogACT} We use LoRA \cite{hu2021loralowrankadaptationlarge} to train the model, with the rank of $32$. We use AdamW \cite{loshchilov2019decoupledweightdecayregularization} with a learning rate of $2\times 10^{-5}$. Training is conducted on $2\times$ H100 for $6000$ steps, with a global batch size of $32$. 
\paragraph{OpenVLA-OFT} We use LoRA \cite{hu2021loralowrankadaptationlarge} to train the model, with the rank of $32$. We use AdamW \cite{loshchilov2019decoupledweightdecayregularization} with a learning rate of $5\times 10^{-4}$. Training is conducted on $2\times$ H100 for $15000$ steps, with a global batch size of $64$. 

For both base models, the coefficients of the training objectives are set as $\alpha_1=0.2, \alpha_2=0.1, \beta=0.1$. For their corresponding benchmarks, the evaluation and the acceleration measurements are all conducted on $1\times$ L40S for each task.

Table~\ref{tab:interval} summarizes the recaching threshold settings and the resulting average recaching intervals.

\begin{table}[ht]
    \centering
    \caption{Recaching threshold settings and the resulting average recaching intervals}
    \label{tab:interval}
    \begin{tabular}{c|ccc}
    \toprule
         & Pick Can & Move Near & Drawer \\
    \midrule
        $\delta_1$ & $0.8$ & $0.4$ & $0.7$ \\
        $\delta_2$ & $0.4$ &$0.3$ &$0.4$ \\
        avg interval of L1 & $9.14$ & $2.23$ & $7.47$  \\
        avg interval of L2 & $2.15$ & $2.05$ & $4.16$ \\
    \bottomrule
    \end{tabular}

\end{table}

\section{Real-World Experiments on Multi-Frame Integration}
\label{app:real_mem}
We summarize the results of real-world experiments on multi-frame integration in Table~\ref{tab:real_mem}. We found that our method outperforms baselines, with an average absolute improvement of $23.3\%$.

\begin{table}[t]
\centering
\caption{Real-world task accuracy (\%) on temporally dependent tasks for multi-frame integration. Each task is evaluated over 20 trials.}
\label{tab:real_mem}
\begin{tabular}{l|ccc|c}
\toprule
Method
& \makecell{Heat\\Vegetables}
& \makecell{Pour\\Twice}
& \makecell{Press\\Buttons}
& Average \\
\midrule
MemoryVLA
& $10.0$
& $25.0$
& $45.0$
& $26.7$ \\
ContextVLA
& $25.0$
& $40.0$
& $40.0$
& $35.0$ \\
\method{} (Ours)
& $\mathbf{35.0}$
& $\mathbf{75.0}$
& $\mathbf{65.0}$
& $\mathbf{58.3}$ \\
\bottomrule
\end{tabular}
\end{table}

\section{Computational and Acceleration Analysis}
\label{app:complexity}
In this section, we provide the full computational analysis of the context length reduction and the acceleration enabled by our approach.

\paragraph{Context length}
Let $N$ be the number of total tokens (including image tokens and text tokens). Let $r$ be the fraction of tokens that are cached, and $N^\prime = (1-r) N$ be the number of recomputed tokens.
% In practice, text tokens are few, usually fewer than 20 tokens in the current VLA benchmarks.

Let $T$ be the number of observations that should be incorporated in the context. The context length with our method reduces from $NT$ to $rN + (1-r)NT=NT-rN(T-1)$.

\paragraph{Static token reuse}
The complexity of the language model backbone in the VLA model mainly comes from two types of modules: the multi-head attention (MHA) layers and the feedforward neural networks. Let $d$ be the dimension of embedding. The FLOPs of the MHA layer are estimated by $4Nd^2+2N^2d$, and the FLOPs of the feedforward neural network are estimated by $2Ndm$, where $m$ is the hidden dimension of the MLP. Therefore, the total FLOPs can be written as $F=4Nd^2+2N^2d+2Ndm$.

With our method, given that the static tokens are cached, only the dynamic parts (including the dynamic image tokens and the language instructions), the projection layers in the MHA layer will reduce to $4N^\prime d^2$, the attention matrix calculation will reduce to $2N^\prime Nd$, and the FLOPs of the feedforward network will reduce to $2N^\prime dm$. The final FLOPs are $F^\prime=4N^\prime  d^2 +2 NN^\prime  d + 2N^\prime  dm$.

% Using the settings and parameters of LLM backbones in prevailing VLAs \cite{kim2024openvla,kim2025fine,li2024cogact} where $d=4096, m=11008, N=276$.
Therefore, the theoretical FLOPs reduction of the LLM backbone \emph{under an idealized setting} could be estimated as
\begin{equation}
% \begin{aligned}
    \frac{F^\prime}{F} = \frac{4N^\prime  d^2 +2 NN^\prime  d + 2N^\prime  dm}{4Nd^2+2N^2d+2Ndm} \\ = (1-r)\frac{4d^2 +2 Nd + 2dm}{4d^2+2Nd+2dm} \\
    =1-r
% \end{aligned}
\end{equation}

The overall reduction will be diluted by the involvement of the vision backbone, the decoding module, and the cache refresh step. However, these components are significantly less computationally intensive than the LLM backbone. Moreover, the introduction of the recache gate incurs only negligible overhead ($\approx 1.27\%$) and has minimal impact on the overall computational complexity. The latency breakdown of each component is summarized in Table~\ref{tab:latency_breakdown}.

\begin{table}[]
    \centering
    \caption{Breakdown of the CUDA latency of each component.}
    \label{tab:latency_breakdown}
    \begin{tabular}{l|cc}
    \toprule
    Component & CUDA Latency (ms) & ratio \\
    \midrule
         Vision Backbone & $246.9$ & $41.1\%$   \\
         Recache Gate & $18.7$ & $3.1\%$ \\
         LLM Backbone & $335.7$ & $55.8\%$ \\
    \midrule 
    Total & $601.2$ & $100.0\%$  \\
    \bottomrule
    \end{tabular}
\end{table}

\section{Latent Representation Drift}
\label{app:drift}

Methods such as \cite{xu2025vla} implicitly assume similarity in the pixel space lead to similarity in the latent space. In this section, we analyze the latent representations during VLA rollouts to show that this assumption is fundamentally flawed.

\begin{figure}
    \centering
    \includegraphics[width=1.0\linewidth]{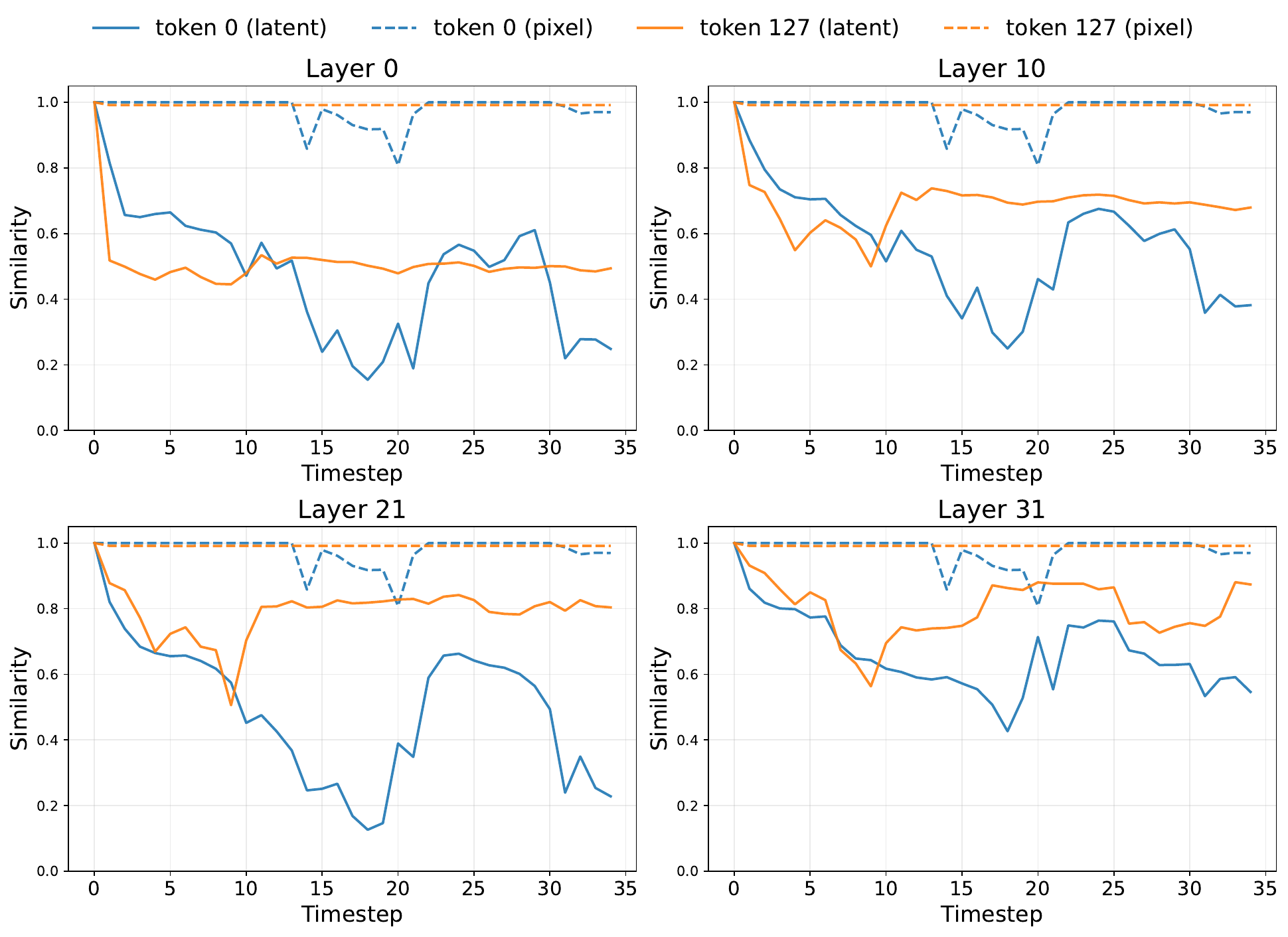}
    \caption{Visually static regions in pixel space exhibit substantial latent representation drift across timesteps. }
    \label{fig:drift}
\end{figure}

As illustrated in Figure~\ref{fig:drift}, we measure cosine similarity between tokens at each timestep and their initial tokens in both pixel and latent space. Visually static tokens still exhibit substantial latent drift — e.g., in layer 0, a pixel-constant token (orange) drops to ~0.5 in latent space after just one timestep, while the blue token with minor pixel changes (steps 15–20) drops from ~0.5 to ~0.2. These results confirm that pixel-level similarity does not imply latent invariance, thus invalidating existing methods and motivating our learned recache mechanism operating in representation space.

\section{Latency Distribution}
\label{app:lat_dist}
\begin{figure}
    \centering
    \includegraphics[width=1.0\linewidth]{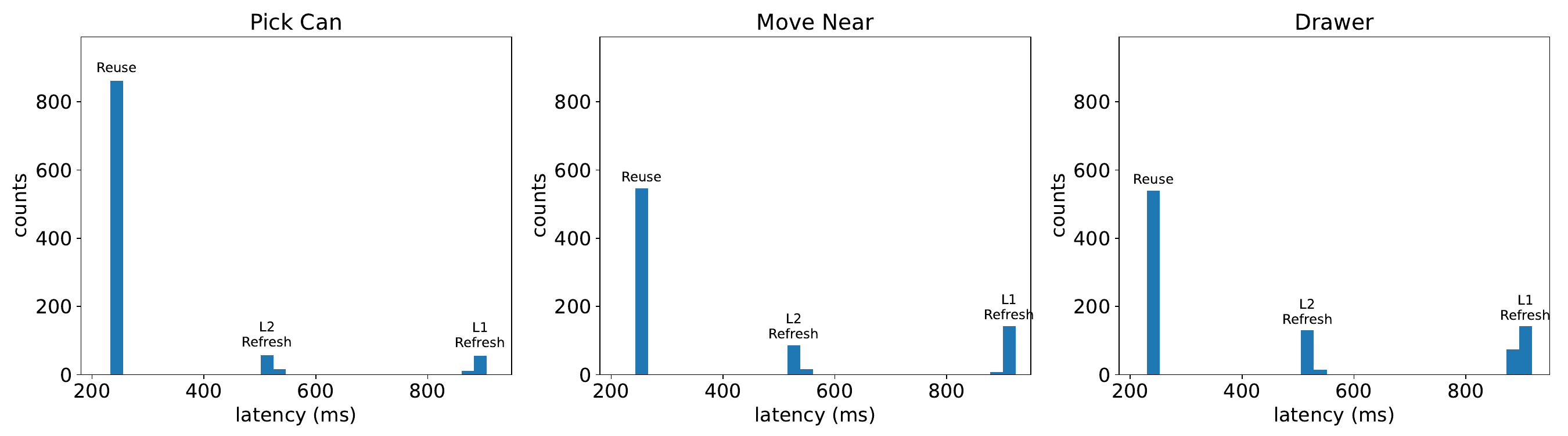}
    \caption{The latency distribution for each step in rollouts. \emph{Reuse} means no refresh needed. \emph{Ln Refresh} means the need to refresh the n-th level static cache.}
    \label{fig:latency}
\end{figure}

We provide the latency distribution during rollout across tasks in Figure~\ref{fig:latency}. On the *Pick Can* task, ~86.0\% of steps fully reuse static tokens across all levels, resulting in low latency (~250 ms). A smaller fraction (~7.4\%) triggers an L2 refresh (~500 ms), while only 6.6\% of steps require an L1 refresh, leading to higher latency (~900 ms).

These results show that high-latency events are infrequent, with the vast majority of steps operating in the efficient reuse regime. Since our goal is to optimize overall rollout efficiency (in addition to enabling multi-frame integration), occasional latency spikes have a negligible impact on end-to-end performance.

\section{Hyperparameter Sensitivity}
In this section, we analyze the sensitivity of crucial hyperparameters.

\begin{figure}
    \centering
    \includegraphics[width=0.8\linewidth]{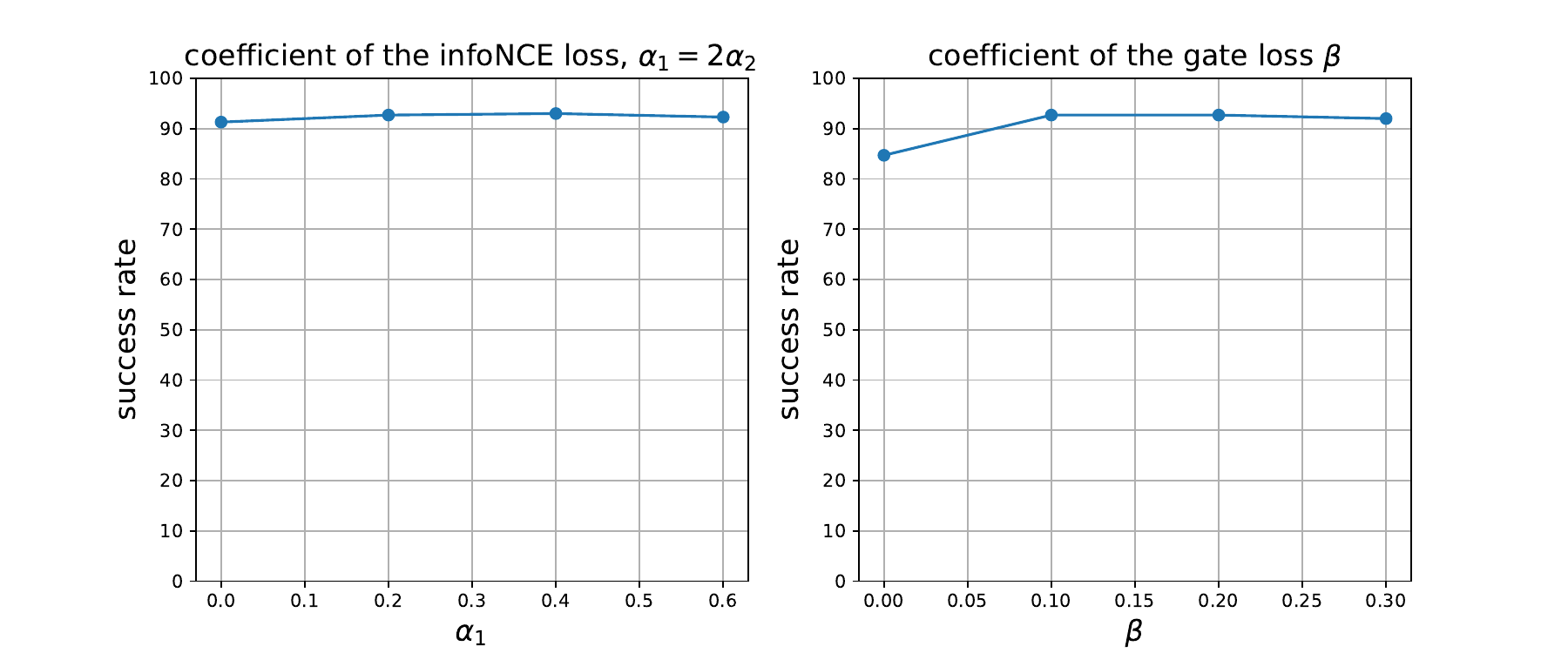}
    \caption{Sensitivity analysis on the coefficients for each loss term. Experiments are conducted on Pick Can (Visual Matching). }
    \label{fig:sensitivity}
\end{figure}

\subsection{Objective Coefficients}
\label{app:sense}

We conduct sensitivity analysis of important hyperparameters in our training objective, i.e., $\alpha$ and $\beta$ in our loss term. The results are plotted in Figure~\ref{fig:sensitivity}. 

For the coefficient $\alpha$ of $\mathcal{L}_\text{infoNCE}$, we find the performance to be stable across a broad range of values.

For the coefficient $\beta$ of $\mathcal{L}_\text{gate}$, we observe that performance is stable once $\beta \geq 0.1$. When $\beta = 0$, the recache gate is not effectively trained and degenerates to random refresh/reuse decisions, leading to significantly degraded performance.

\subsection{Static Ratio}
\label{app:static_ratio}
We consider the static ratio as a tunable hyperparameter that can be adapted to different environments. For more dynamic settings, it can be adjusted by (1) reducing the ratio during training to allocate more capacity to dynamic tokens, or (2) lowering the recache threshold for more frequent cache refreshing at test time.

\begin{figure}
    \centering
    \includegraphics[width=0.5\linewidth]{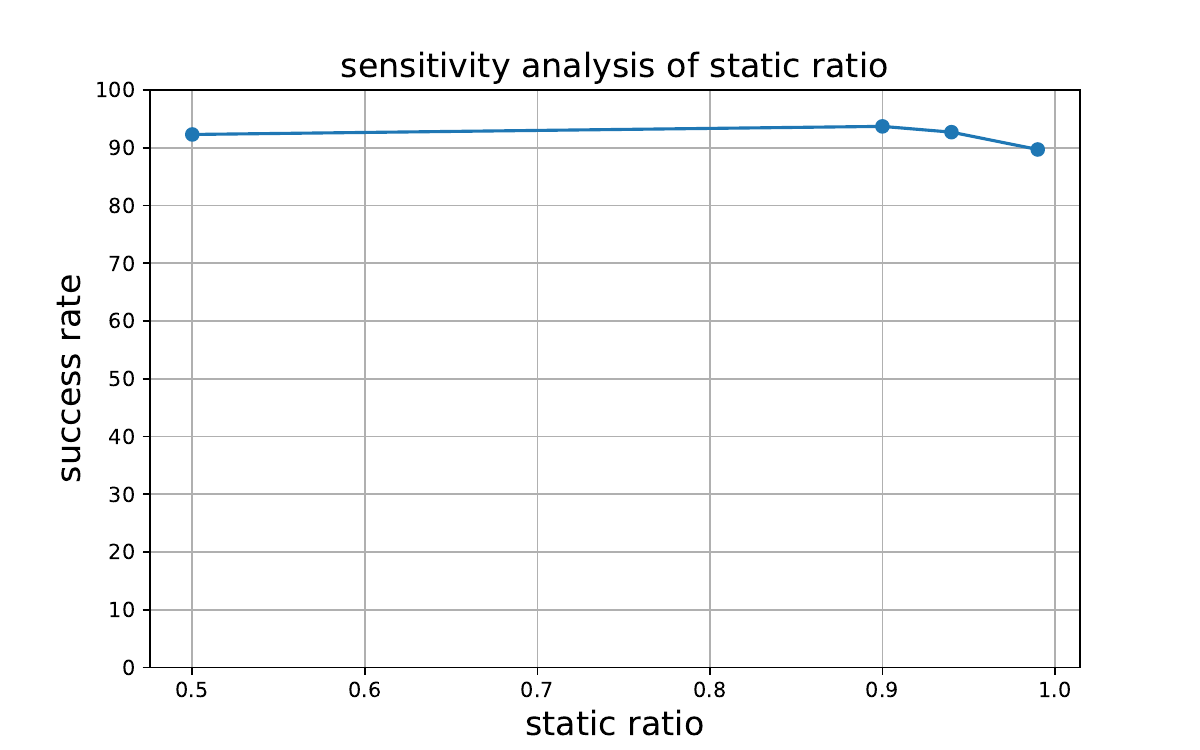}
    \caption{Sensitivity analysis of the static ratio. Experiments are conducted on Pick Can (Visual Matching).}
    \label{fig:static}
\end{figure}

Our sensitivity analysis in Figure~\ref{fig:static} shows stable performance across a broad range of ratios, with degradation only at extreme values.

% \begin{table}[]
%     \centering
%     \caption{Success rate (\%) on wrist image. Our model still achieves acceleration without the sacrifice of performance.}
%     \label{tab:real}
%     \begin{tabular}{l|c|c}
%     \toprule
%     Method & LIBERO-10 (wrist) & FLOPs \\
%     \midrule
%         OpenVLA-OFT &  $88.8$ & $100.0\%$ \\
%         + SD-VLA & $\mathbf{89.0}$ & $\mathbf{72.3\%}$\\
%     \bottomrule
%     \end{tabular}
% \end{table}

% \begin{table}[]
%     \centering
%     \caption{Success rate (\%) on LIBERO benchmark with $\pi_{0.5}$ \cite{intelligence2025pi05visionlanguageactionmodelopenworld}.}
%     \label{tab:real}
%     \begin{tabular}{l|cc|c}
%     \toprule
%     Method & Long & Spatial & FLOPs \\
%     \midrule
%         $\pi_{0.5}$ & $94.0\%$ & $96.8\%$ & $100.0\%$ \\
%         + SD-VLA & $\mathbf{94.2}\%$ & $\mathbf{97.8\%}$ & $\mathbf{48.9\%}$\\
%     \bottomrule
%     \end{tabular}
% \end{table}

\section{Temporal Evolution of Static-Dynamic Assignments}
\label{app:time_evo}

\begin{figure}
    \centering
    \includegraphics[width=\linewidth]{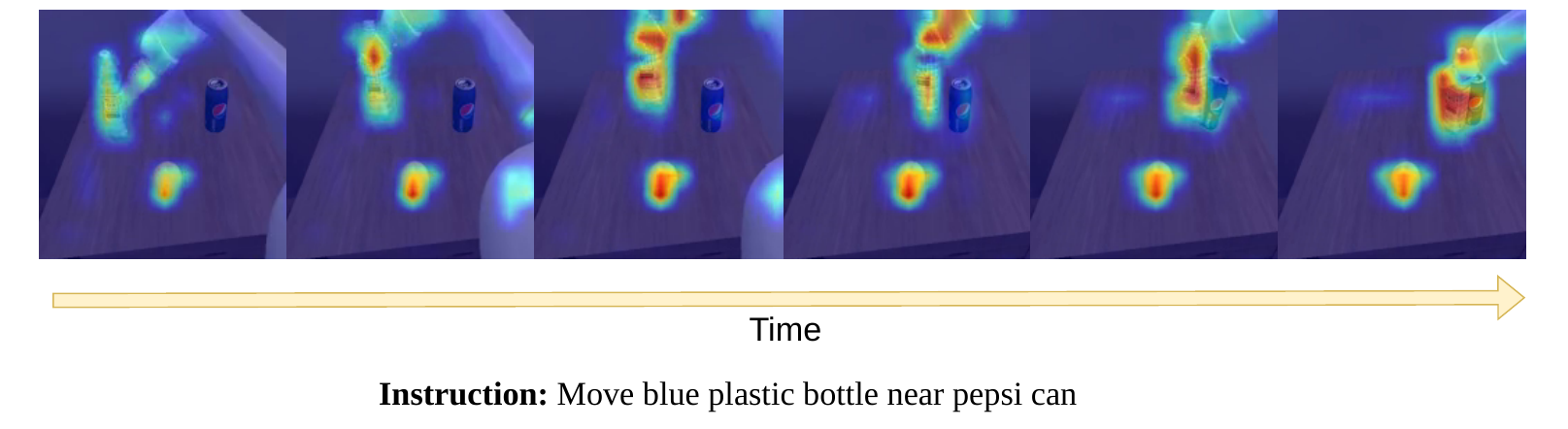}
    \caption{An illustration that the model misclassifies objects and later corrects such errors.}
    \label{fig:correctify}
\end{figure}

Below, we show a representative trajectory illustrating how \method{} adapts its static-dynamic assignments over time. As shown in Figure~\ref{fig:correctify}, the Pepsi can is initially treated as static because it is not intended to be manipulated. However, when unexpected contact occurs, the model detects the resulting change and reclassifies the can as dynamic, thereby triggering a cache refresh. In some cases, objects such as the orange may be treated as dynamic even though they remain unchanged. This does not affect task correctness, but only introduces a small amount of additional computational overhead.

% \begin{figure}
%     \centering
%     \includegraphics[width=\linewidth]{figs/fail.pdf}
%     \caption{Real-world failure cases of the VLA base model (OpenVLA-OFT \cite{kim2025fine})}
%     \label{fig:fail}
% \end{figure}

\section{Impact Statement}
This paper presents work whose goal is to advance the field of Machine Learning. There are many potential societal consequences of our work, none of which we feel must be specifically highlighted here.

\section{Declaration of LLM Usage}
We used large language models solely for editing and polishing the manuscript (e.g., grammar, phrasing, and typo correction). LLMs were not involved in the research ideation, methodology, experiments, or analysis, and do not constitute any original or non-standard component of this work.
%%%%%%%%%%%%%%%%%%%%%%%%%%%%%%%%%%%%%%%%%%%%%%%%%%%%%%%%%%%%

\newpage
\section*{NeurIPS Paper Checklist}

\begin{enumerate}

\item {\bf Claims}
    \item[] Question: Do the main claims made in the abstract and introduction accurately reflect the paper's contributions and scope?
    \item[] Answer: \answerYes{} % Replace by \answerYes{}, \answerNo{}, or \answerNA{}.
    \item[] Justification: The three contributions stated in the introduction are each developed in the Method section and validated in the Experiments section, with scope explicitly stated.

\item {\bf Limitations}
    \item[] Question: Does the paper discuss the limitations of the work performed by the authors?
    \item[] Answer: \answerYes{} % Replace by \answerYes{}, \answerNo{}, or \answerNA{}.
    \item[] Justification: The paper has included a limitation paragraph in the last section.

\item {\bf Theory assumptions and proofs}
    \item[] Question: For each theoretical result, does the paper provide the full set of assumptions and a complete (and correct) proof?
    \item[] Answer: \answerYes{}
    \item[] Justification: The paper provides computational analysis in the main text and its derivation in the appendix.

    \item {\bf Experimental result reproducibility}
    \item[] Question: Does the paper fully disclose all the information needed to reproduce the main experimental results of the paper to the extent that it affects the main claims and/or conclusions of the paper (regardless of whether the code and data are provided or not)?
    \item[] Answer: \answerYes{} % Replace by \answerYes{}, \answerNo{}, or \answerNA{}.
    \item[] Justification: The paper describes the experimental setting in the main text and provides the details in the appendix. The code has also been provided in the supplemental material.

\item {\bf Open access to data and code}
    \item[] Question: Does the paper provide open access to the data and code, with sufficient instructions to faithfully reproduce the main experimental results, as described in supplemental material?
    \item[] Answer: \answerYes{} % Replace by \answerYes{}, \answerNo{}, or \answerNA{}.
    \item[] Justification: The code and the instructions have been provided in the supplemental material, and the data is publicly available for the simulation experiments.

\item {\bf Experimental setting/details}
    \item[] Question: Does the paper specify all the training and test details (e.g., data splits, hyperparameters, how they were chosen, type of optimizer) necessary to understand the results?
    \item[] Answer: \answerYes{} % Replace by \answerYes{}, \answerNo{}, or \answerNA{}.
    \item[] Justification: The paper describes the experimental setting in the main text and provides the details in the appendix.

\item {\bf Experiment statistical significance}
    \item[] Question: Does the paper report error bars suitably and correctly defined or other appropriate information about the statistical significance of the experiments?
    \item[] Answer: \answerNo{} % Replace by \answerYes{}, \answerNo{}, or \answerNA{}.
    \item[] Justification: We do not report error bars across training runs due to the prohibitive computational cost of training VLA models—each reported number can require multiple days on H100-class GPUs—which makes multi-seed training infeasible within the resource budget of this work. This is consistent with standard practice in prior VLA literature. Each result is instead aggregated over multiple evaluation episodes per task to mitigate evaluation variance.

\item {\bf Experiments compute resources}
    \item[] Question: For each experiment, does the paper provide sufficient information on the computer resources (type of compute workers, memory, time of execution) needed to reproduce the experiments?
    \item[] Answer: \answerYes{} % Replace by \answerYes{}, \answerNo{}, or \answerNA{}.
    \item[] Justification: The paper describes the compute resources used for the experiments in in the appendix.
    
\item {\bf Code of ethics}
    \item[] Question: Does the research conducted in the paper conform, in every respect, with the NeurIPS Code of Ethics \url{https://neurips.cc/public/EthicsGuidelines}?
    \item[] Answer: \answerYes{} % Replace by \answerYes{}, \answerNo{}, or \answerNA{}.
    \item[] Justification: The paper advances general machine learning research and does not raise any ethical concerns. The authors have reviewed the NeurIPS Code of Ethics and confirm that their research conforms to it.

\item {\bf Broader impacts}
    \item[] Question: Does the paper discuss both potential positive societal impacts and negative societal impacts of the work performed?
    \item[] Answer: \answerYes{} % Replace by \answerYes{}, \answerNo{}, or \answerNA{}.
    \item[] Justification: The author has included an impact statement section in the appendix.
    
\item {\bf Safeguards}
    \item[] Question: Does the paper describe safeguards that have been put in place for responsible release of data or models that have a high risk for misuse (e.g., pre-trained language models, image generators, or scraped datasets)?
    \item[] Answer: \answerNA{} % Replace by \answerYes{}, \answerNo{}, or \answerNA{}.
    \item[] Justification: The released assets are a robot manipulation model and a simulation benchmark (LIBERO-Memory), which do not generate harmful content nor involve scraped or sensitive data, and therefore pose no high risk of misuse.

\item {\bf Licenses for existing assets}
    \item[] Question: Are the creators or original owners of assets (e.g., code, data, models), used in the paper, properly credited and are the license and terms of use explicitly mentioned and properly respected?
    \item[] Answer: \answerYes{} % Replace by \answerYes{}, \answerNo{}, or \answerNA{}.
    \item[] Justification: All pre-trained base models, datasets, and simulation frameworks are properly cited and used in accordance with their original licenses.

\item {\bf New assets}
    \item[] Question: Are new assets introduced in the paper well documented and is the documentation provided alongside the assets?
    \item[] Answer: \answerYes{} % Replace by \answerYes{}, \answerNo{}, or \answerNA{}.
    \item[] Justification: We introduce the benchmark and the codebase. Both are documented in the appendix (dataset construction, BDDL scene description, oracle demonstrations, and implementation details) and provided in the supplemental material.

\item {\bf Crowdsourcing and research with human subjects}
    \item[] Question: For crowdsourcing experiments and research with human subjects, does the paper include the full text of instructions given to participants and screenshots, if applicable, as well as details about compensation (if any)? 
    \item[] Answer: \answerNA{} % Replace by \answerYes{}, \answerNo{}, or \answerNA{}.
    \item[] Justification: The paper does not involve crowdsourcing. The human-performance reference in the appendix was conducted by two trained volunteers, which only offers as an additional context for understanding the experiments.

\item {\bf Institutional review board (IRB) approvals or equivalent for research with human subjects}
    \item[] Question: Does the paper describe potential risks incurred by study participants, whether such risks were disclosed to the subjects, and whether Institutional Review Board (IRB) approvals (or an equivalent approval/review based on the requirements of your country or institution) were obtained?
    \item[] Answer: \answerNA{} % Replace by \answerYes{}, \answerNo{}, or \answerNA{}.
    \item[] Justification: The paper does not involve study on human subjects.

\item {\bf Declaration of LLM usage}
    \item[] Question: Does the paper describe the usage of LLMs if it is an important, original, or non-standard component of the core methods in this research? Note that if the LLM is used only for writing, editing, or formatting purposes and does \emph{not} impact the core methodology, scientific rigor, or originality of the research, declaration is not required.
    %this research?
    \item[] Answer: \answerYes{} % Replace by \answerYes{}, \answerNo{}, or \answerNA{}.
    \item[] Justification: The paper includes a declaration in the appendix stating that LLMs were used only for editing and polishing the manuscript, with no involvement in the core methodology.

\end{enumerate}

\end{document}